\documentclass[10pt,twocolumn,letterpaper]{article}

\usepackage[pagenumbers]{cvpr} % To force page numbers, e.g. for an arXiv version

\usepackage{graphicx}
\usepackage{amsmath}
\usepackage{amssymb}
\usepackage{booktabs}

\usepackage[pagebackref,breaklinks,colorlinks]{hyperref}

\usepackage[capitalize]{cleveref}
\crefname{section}{Sec.}{Secs.}
\Crefname{section}{Section}{Sections}
\Crefname{table}{Table}{Tables}
\crefname{table}{Tab.}{Tabs.}

\usepackage{amsmath,amsfonts,bm}

\def\eqref#1{equation~\ref{#1}}
\def\1{\bm{1}}

\DeclareMathAlphabet{\mathsfit}{\encodingdefault}{\sfdefault}{m}{sl}
\SetMathAlphabet{\mathsfit}{bold}{\encodingdefault}{\sfdefault}{bx}{n}

\usepackage{booktabs}
\usepackage{makecell}
\usepackage{fontawesome5}
\usepackage{lipsum}
\usepackage{comment}
\usepackage{multirow}
\usepackage{wrapfig}
\usepackage{algorithm}
\usepackage{algorithmic}
\usepackage{subfigure}

\usepackage{graphicx}
\usepackage{color}

\usepackage[english]{babel}
\usepackage{amsthm}
\usepackage{xcolor}
\definecolor{mygray}{gray}{0.7}
\definecolor{darkgreen}{rgb}{0,0.6,0}
\newcommand{\gray}[1]{{\color{mygray}{#1}}}

\newcommand{\green}[1]{{\color{darkgreen}{#1}}}

\newcommand{\A}{{\mathcal{A}}}

\newcommand{\D}{\mathcal{D}}

\renewcommand{\L}{\mathcal{L}}
\newcommand{\N}{\mathcal{N}}
\renewcommand{\O}{\mathcal{O}}
\renewcommand{\P}{\mathcal{P}}

\newcommand{\U}{\mathcal{U}}

\renewcommand{\a}{{\bm{a}}}

\def\eg{\emph{e.g.}} 
\def\ie{\emph{i.e.}}

\def\wrt{{w.r.t.\ }}

\begin{document}

\title{GreedyNASv2: Greedier Search with a Greedy Path Filter}

\author{%
	Tao Huang${}^{1,2}$ \quad Shan You$^{1,3}$\thanks{Correspondence to: Shan You $<$\texttt{youshan@sensetime.com}$>$.}  \quad Fei Wang$^4$  
	\\
	Chen Qian$^1$ \quad Changshui Zhang$^{3}$ \quad Xiaogang Wang$^{1,5}$ \quad Chang Xu$^2$\\
	%~\\
	\normalsize $^1$SenseTime Research \quad
	\normalsize $^2$School of Computer Science, Faculty of Engineering, The University of Sydney \\
	\normalsize $^3$Department of Automation, Tsinghua University,
    \normalsize Institute for Artificial Intelligence, Tsinghua University (THUAI), \\
    \normalsize Beijing National Research Center for Information Science and Technology (BNRist)\\
	\normalsize $^4$University of Science and Technology of China \quad $^5$The Chinese University of Hong Kong\\ 
}

\begin{comment}
\author{First Author\\
Institution1\\
Institution1 address\\
{\tt\small firstauthor@i1.org}
\and
Second Author\\
Institution2\\
First line of institution2 address\\
{\tt\small secondauthor@i2.org}
}
\end{comment}
\maketitle

%%%%%%%%% ABSTRACT
\begin{abstract}
Training a good supernet in one-shot NAS methods is difficult since the search space is usually considerably huge (\eg, $13^{21}$). In order to enhance the supernet's evaluation ability, one greedy strategy is to sample good paths, and let the supernet lean towards the good ones and ease its evaluation burden as a result. However, in practice the search can be still quite inefficient since the identification of good paths is not accurate enough and sampled paths still scatter around the whole search space. In this paper, we leverage an explicit path filter to capture the characteristics of paths and directly filter those weak ones, so that the search can be thus implemented on the shrunk space more greedily and efficiently. Concretely, based on the fact that good paths are much less than the weak ones in the space, we argue that the label of ``weak paths" will be more confident and reliable than that of ``good paths" in multi-path sampling. In this way, we thus cast the training of path filter in the positive and unlabeled (PU) learning paradigm, and also encourage a \textit{path embedding} as better path/operation representation to enhance the identification capacity of the learned filter. By dint of this embedding, we can further shrink the search space by aggregating similar operations with similar embeddings, and the search can be more efficient and accurate. Extensive experiments validate the effectiveness of the proposed method GreedyNASv2. For example, our obtained GreedyNASv2-L achieves $81.1\%$ Top-1 accuracy on ImageNet dataset, significantly outperforming the ResNet-50 strong baselines.
\end{abstract}

%%%%%%%%% BODY TEXT
\section{Introduction} \label{sec:intro}

Neural architecture search (NAS) aims to boost the performance of deep learning by seeking an optimal architecture in the given space, and it has achieved significant improvements in the sight of applications, such as image classification \cite{tan2019efficientnet,guo2020single,you2020greedynas,su2021vision,yang2020ista} and object detection \cite{chen2019detnas,guo2020hit}. One-shot NAS \cite{guo2020single,you2020greedynas,su2021prioritized,yang2021towards,huang2020explicitly,su2021bcnet,su2020locally} stands out from the literature of NAS for the sake of its decent searching efficiency. Instead of exhaustively training each possible architecture, one-shot NAS fulfills the searching in an only one-shot trial, where a supernet is leveraged to embody all candidate architectures (\ie, paths). Each path can be parameterized by the corresponding weights within the supernet, and thus gets trained, evaluated, and ranked. 
Typical uniform sampling (SPOS) \cite{guo2020single} is usually adopted to train the supernet because of the feasible single-path memory consumption and being friendly to large-scale datasets. 

\begin{figure}[t]
    \centering
    \includegraphics[width=1\linewidth]{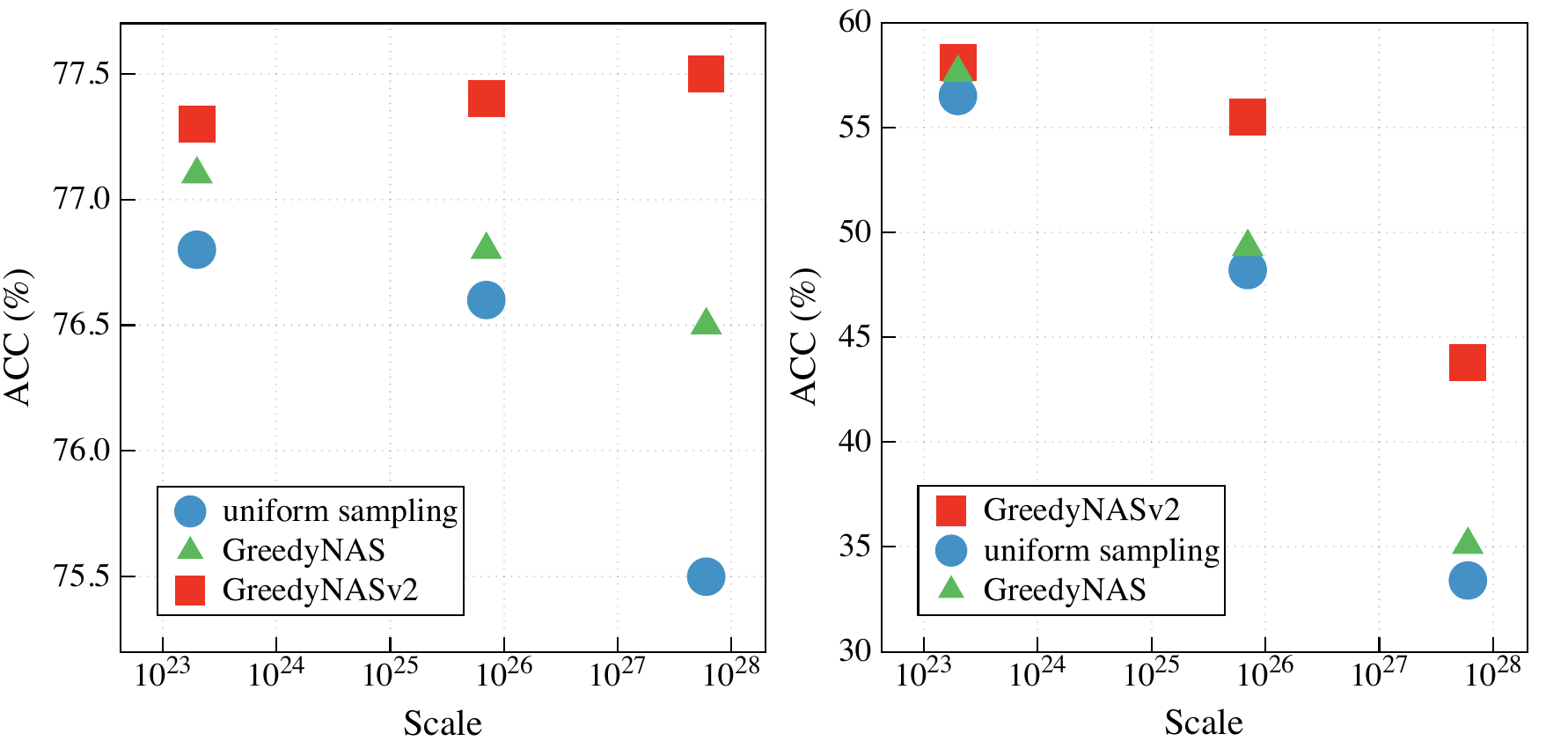}
    %\fbox{\rule[-.5cm]{0cm}{4cm} \rule[-.5cm]{4cm}{0cm}}
    \vspace{-6mm}
    \caption{Performance of searched architectures \wrt different scales of search space. Left: retraining accuracies of models searched by GreedyNASv2 and baselines. Right: validation accuracies of searched models on supernets.}
    \label{fig:figure1}
    \vspace{-4mm}
\end{figure}

\begin{figure*}[t]
    \centering
    \includegraphics[width=0.95\linewidth]{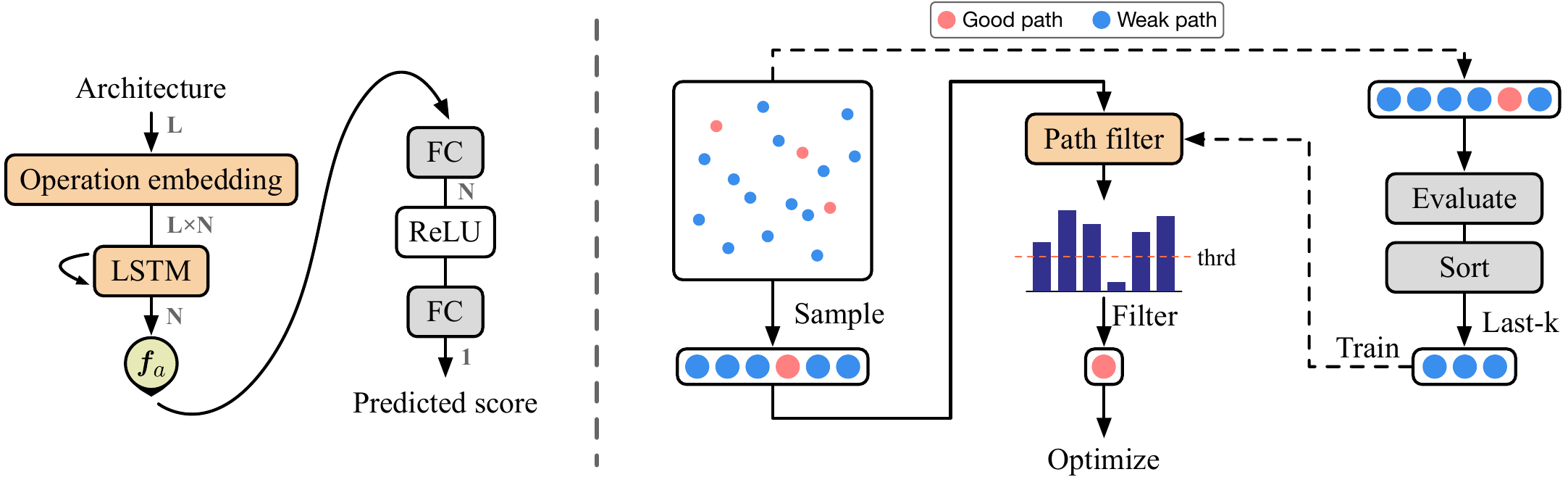}
    \caption{Left: the architecture of our path filter. Right: diagram of supernet training in GreedyNASv2. In GreedyNASv2, we adopt a path filter to filter weak paths from the uniformly-sampled paths, and the remained potentially-good paths are then used for optimization. The path filter is trained using weak paths identified by a validation set and unlabeled paths.}
    \label{fig:figure2}
    \vspace{-4mm}
\end{figure*}

The architecture search space in NAS could be considerably huge (\eg, $13^{21}$). Equally treating different paths and uniformly sampling them from the supernet could lead to inappropriate training of the supernet, as the weak paths would disturb the highly-shared weights. Various sampling strategies have thus been proposed to address this issue, such as fair sampling \cite{chu2021fairnas} and Monte-Carlo tree search \cite{su2021prioritized}. We are particularly interested in the strategy of multi-path sampling with rejection by GreedyNAS \cite{you2020greedynas}, which identifies good paths from weak paths and then only greedily updates those good ones; it is easy to implement and more suitable for large search spaces among these methods. Working on the whole search space, GreedyNAS has to safely allocate only a medium level of partition of good paths (\eg, only 5 out of 10) to ensure a high probability of sampled paths being good. But the search will become infeasible and limited if the search space grows larger with more operation choices. Besides, GreedyNAS needs to maintain a candidate pool to recycle paths, which limits the number of stored paths, and many \textit{elite paths} could be missed. 

In this paper, we propose GreedyNASv2 to power the multi-path sampler with explicit search space shrinkage for one-shot NAS, which targets on a greedier search space with an tiny (\eg, only 1\%) proportion of paths treated as ``good paths''. Since good paths are usually much less than weak paths, the probability of picking out a good path by a multi-path sampler could be smaller than that of sampling weak paths. If weak paths can be captured with confidence, we can easily screen them out from the searching space and execute a greedier search on the shrunk space. By doing so, the supernet only needs to focus on evaluating those not too bad paths (potentially-good paths), which benefits the overall searching performance and efficiency simultaneously.

The key is then to learn a path filter to identify those weak paths from the entire architecture search space. Though it is hard to find a good path, we can have high confidence about weak paths in the multi-path sampling. These identified weak paths with confidence can be regarded as positive examples to be thrown away. As a precaution, the remaining paths in the search space are taken as unlabeled examples, as they may contain both weak paths (positive examples to be thrown away) and good paths (negative examples not to be thrown away). The learning of this path filter can thus be formulated as the Positive-Unlabeled (PU) learning problem \cite{letouzey2000learning, elkan2008learning}. Once the path filter has been well trained,  a given new path can be efficiently predicted to specify whether it is weak or not. A path embedding is also learned with the path filter to encode the path as a better path representation. Since the path embedding is learned in the weak/good sense, if two operations have similar embeddings, it means that both operations have similar or even the same impact on discriminating paths, and they can be thus merged. This enables a greedy shrinkage of operations, which is expected to work together with the path shrinkage to boost the searching performance and efficiency further. 

We conduct extensive experiments on ImageNet dataset to validate the effectiveness of our proposed GreedyNASv2. Compared to the baseline methods SPOS and GreedyNAS, our proposed method achieves better performance with less search cost. To further investigate our superiority, we even search on a larger space, which has $\sim 10^4\times$ architectures compared to the commonly-used MobileNetV2-SE search space, and the results show that our searched model outperforms state-of-the-art NAS models. The performance on different scales of search spaces are illustrated in Figure~\ref{fig:figure1}. Besides, we also compare the searching performance on a recent benchmark NAS-Bench-Macro \cite{su2021prioritized} for one-shot NAS. Ablation studies show that our GreedyNASv2 effectively samples better architectures than uniform sampling and the multi-path sampler in GreedyNAS.

\section{Related Work}

\subsection{NAS with search space shrinkage}
\textbf{Path-level shrinkage.} To obtain a path-level shrinkage on search space, GreedyNAS~\cite{you2020greedynas} proposes a candidate pool to store those evaluated good paths and samples from it using an exploration-exploitation strategy. MCT-NAS~\cite{su2021prioritized} proposes to sample architectures with the guidance of Monte-Carlo tree search; hence the good paths can be sampled with better exploration and exploitation balance. However, the limited size (\eg, 1000) of candidate pool in GreedyNAS is too aggressive to train the elite paths with enough diversity, and the exponentially-increased Monte-Carlo tree makes the MCT-NAS difficult to scale to larger search spaces.

\textbf{Operation-level shrinkage.} Operation-level shrinkage is also an effective way to reduce both training parameters and the size of search space. Some methods~\cite{hu2020angle,shen2020bs} design importance metrics to identify the good operations and drop the weak ones. For example, ABS~\cite{hu2020angle} measures the importance of each operation using the angle between its trained weights and initialized weights; BS-NAS~\cite{shen2020bs} proposes a channel-level importance metric by measuring a number of architectures on the validation dataset. However, these methods only consider operation-level statistics, while for each specific architecture, the preferences of operations may be different. On the other hand, NSENet~\cite{ci2021evolving} proposes to learn the importance using additional learnable indicators after each operation, which is learned by simulating the gradients of binary-selected architectures. However, this simulation of gradients introduces approximation errors and also increases memory consumption. 

In this paper, we perform both path-level and operation-level shrinkage using a path filter. The path filter is constructed by a binary classifier, which efficiently filters the weak paths and generalizes well to the whole search space; hence we can filter the weak paths more greedily. Furthermore, we can perform operation-level shrinkage without extra costs by measuring the learned operation embeddings in the path filter. This operation merging strategy holds naturally since the operations with similar embeddings would have similar predictions and thus similar performance.

\subsection{Positive-unlabeled learning}
Positive-Unlabeled (PU) classification is a problem of training a binary classifier from only positive and unlabeled data \cite{letouzey2000learning, elkan2008learning}. 
\begin{comment}
In NAS, it only has a very small portion of good paths in the search space, so that we have a small confidence on labeling good paths while the confidence on weak paths is fairly large. This naturally formulates a PU problem: \textit{with only labeled reliable weak paths and unlabeled paths, how can we learn a predictor to distinguish the good and weak architectures?} Luckily, this problem is widely discussed in PU learning and many effective methods~\cite{du2015convex,kiryo2017positive,chen2020vpu} are proposed to train a good binary classifier. 
\end{comment}
Many effective methods~\cite{du2015convex,kiryo2017positive,chen2020vpu} are proposed to train a good binary classifier in PU learning. Specifically, uPU~\cite{du2015convex} rewrites the classification risk in terms of the distributions over positive and unlabeled samples, and obtains an unbiased estimator of the risk without negative samples. To overcome the overfitting problem in uPU, a non-negative risk estimator is proposed in nnPU\cite{kiryo2017positive}. One recent approach VPU~\cite{chen2020vpu} proposes a variational principle for PU learning without involving class prior estimation or any other intermediate estimation problems. In this paper, we implement VPU to learn our path filter.

\section{Revisiting Multi-path Sampler}
In single path one-shot NAS~\cite{guo2020single, you2020greedynas, su2021prioritized}, the search space is treated as an over-parameterized supernet $\N$, in which the searching layers are stacked sequentially, and each layer is required to select one operation from candidate operations. Assume the supernet has $L$ layers and $N$ candidate operations $\O=\{o_i\}, \forall i = 1,2,...,N$, then each architecture can be represented by a tuple with size $L$, \ie, $\bm{a}=(o^{(1)}, o^{(2)}, ..., o^{(L)})$, where $o^{(j)} \in \O$, $\forall j = 1,2,...,L$. As a result, the search space $\A$ is of size $|\A| = N^L$. With a pre-defined supernet, the NAS procedure is split into two stages: supernet training and search. During training, the supernet is optimized by alternately sampling paths and updating their corresponding weights. Thereafter, the optimal path can be determined as the one with the highest accuracy on a hold-out validation set. 

\begin{figure}[t]
    \centering
    \includegraphics[width=1\linewidth]{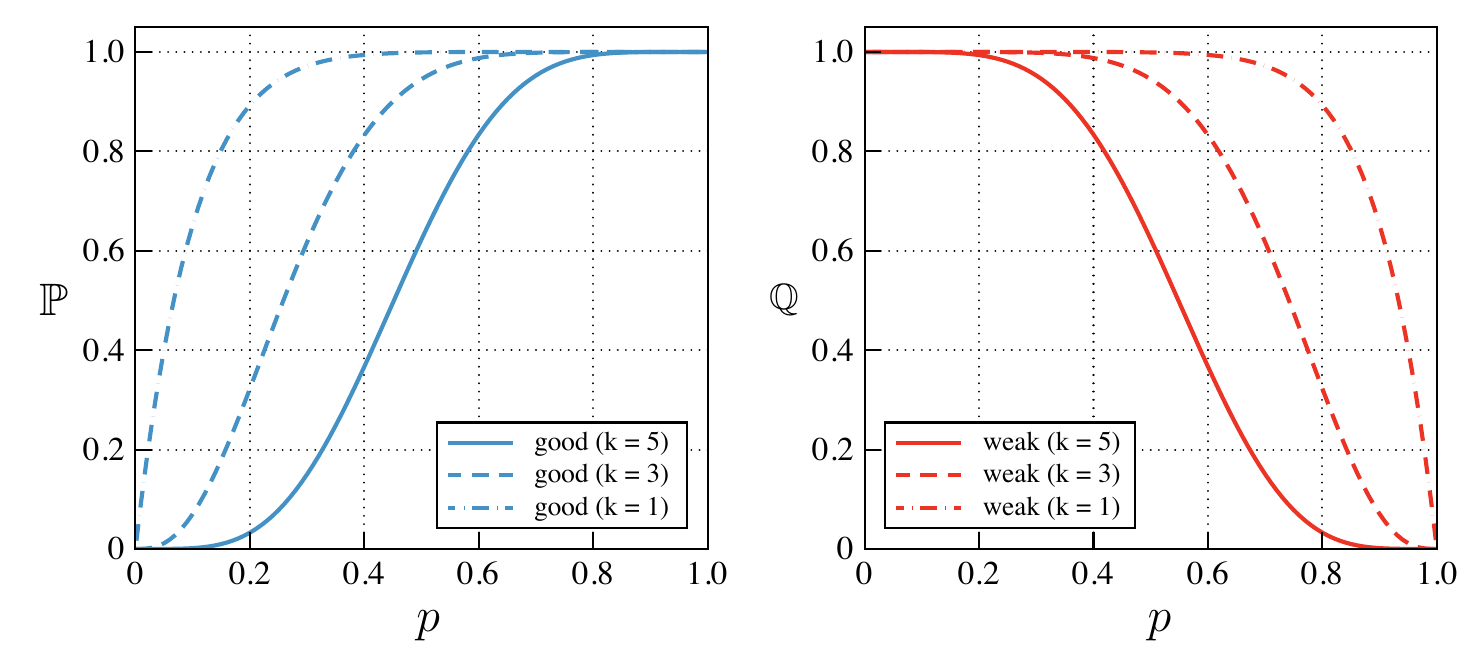}
    \vspace{-6mm}
    \caption{Confidence $\mathbb{P}$ ($\mathbb{Q}$) of sampling at least $k$ good (weak) paths out of $m=10$ paths with proportion $p$ of good paths.}
    \vspace{-2mm}
    \label{fig:confidence}
\end{figure}

\begin{figure*}[t]
    \centering
    \includegraphics[width=0.95\linewidth]{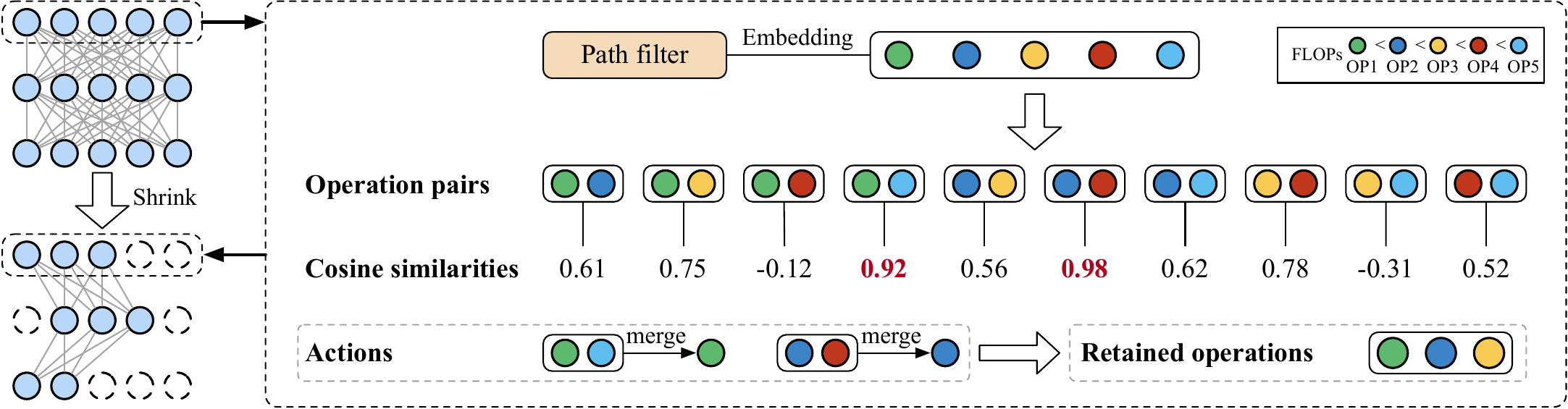}
    \caption{Overview of the proposed operation shrinkage method. We evaluate the cosine similarities of operation pairs in each layer using the learned operation embeddings. Then, we merge the similar operation pair to one operation with smaller FLOPs.}
    \label{fig:merge}
\end{figure*}

Although the supernet shares weights with all architectures, it still has $\sim N\times$ parameters than a common path. For example, the benchmark MobileNetV2-SE search space has $13$ operations and $46$M parameters for supernet, while a path only has $\sim 5$M parameters. With such a large supernet, it is harsh to optimize all the architectures well and evaluate them accurately. Therefore, instead of sampling paths uniformly \cite{guo2020single}, GreedyNAS~\cite{you2020greedynas} proposes a multi-path sampling with rejection to greedily sample those potentially-good paths; hence the training efficiency and performance can be boosted.

\subsection{Frustrating sampling good paths}

In the multi-path sampler, $m$ paths are sampled at once, then evaluated and ranked by a small validation set. According to the Theorem~\ref{theorem1} below, we can thus safely treat the Top-$k$ paths as good ones, and train the supernet greedily by just updating these $k$ paths. 
\newtheorem{theorem}{Theorem}
\begin{theorem}[multi-path sampling \cite{you2020greedynas}] \label{theorem1}
If $m$ paths are sampled uniformly i.i.d from $\A$, and the proportion of good paths in the search space is $p$, then it holds that at least $k(k\le m)$ paths are good paths with confidence
\begin{equation}
  \mathbb{P} :=   \sum_{j=k}^m\mathbb{C}_m^j p^j (1-p)^{m-j} .
\end{equation}
\end{theorem}
To ensure a high confidence $\mathbb{P}$ of sampling good paths, GreedyNAS only has to assume a medium level of good path proportion (\ie, large $p$). For example, $\mathbb{P}$ comes to $83.38\%$ with $p=0.6$ when we leave $k=5$ paths as good from the sampled $m=10$ paths.

However, this is not enough. Since we target at the optimal architecture, the candidate \textit{elite} paths are supposed to be way less than the weak ones, which means that $p\ll 0.5$ holds naturally, and thus we can have an actually shrunk space to boost the searching. Frustratingly, the confidence will degrade according; for example, with $p=0.1$, the previous confidence $\mathbb{P}$ will be only $0.16\%$. Though GreedyNAS leverages a candidate pool to recycle paths, many \textit{elite paths} could be missed since it heavily relies on the limited number of stored paths (\eg, 1000).

\subsection{Turning tables with weak paths}
Sampling good paths is frustratingly ineffective in a more greedy space since the confidence collapses as a failure. In contrast, since the good prior $p$ is low, the search space will be glutted with weak paths, and the probability of sampling a weak path $q:= 1-p$ is thus large accordingly. Similarly, based on Theorem \ref{theorem1}, in multi-path sampling the confidence of sampled weak paths goes even larger, denoted as $\mathbb{Q}:= \sum_{j=k}^m\mathbb{C}_m^j q^j (1-q)^{m-j}$. For example, with $q=0.9$ ($p=0.1$), the probability of sampling at least $5$ weak paths out of $10$ is very high ($\mathbb{Q}=99.99\%$). See Figure \ref{fig:confidence} for more details. 

Now the tables have been turned. If we can sample weak paths with high confidence, we can easily rule them out from the entire search space and implement a more greedily searching on the shrunk space, thus boosting the searching performance as well. Then the question goes to: how can we leverage the sampled weak paths to identify a shrunk space composed of good paths? Intuitively, we encourage learning a path filter to encode the characteristics of sampled weak paths and identify the label of a given new path.

Nevertheless, during multi-path sampling, we only have confidence about weak paths, \textit{can we still learn a discriminative path filter as a decent binary classifier to predict labels of paths?} The answer is affirmative; in the sequel, we will cast the learning as a typical Positive and Unlabeled (PU) problem. 

\section{Greedier Sampling with a Path Filter}
\begin{comment}
In this paper, we propose to achieve greedier sampling with a path filter, which is a binary classifier to predict whether a path is weak or not. We greedily filter those paths with weak predictions and sample the remained good ones for optimization. Since the path filter predicts paths much faster than evaluating paths in multi-path sampling, we can efficiently predict a batch of paths one time and filter a large percentage of paths for greedy training.
\end{comment}

After multi-path sampling, we now have confident weak paths; nevertheless, remaining paths are difficult to specify whether they are weak or good, since the corresponding confidence will be low. As a precaution, we regard the remaining paths (together with unsampled paths) as unlabeled examples, as they may contain both weak and good paths. 

\subsection{Learning path filter as PU prediction}
Here we want to learn a path filter with the identified weak paths (positive examples) and remaining paths (unlabeled examples). Formally, let us first consider a binary classification problem where the architectures $\a\in\A$ and class labels $y\in\{-1,+1\}$ are distributed according to a joint distribution $\mathbb{D}(\a,y)$, and the paths with positive label $+1$ denote weak paths to be discarded. In GreedyNASv2, we have positive dataset $\P = \{\a_1, ..., \a_M\}$ and unlabeled dataset $\U = \{\a_{M+1},...,\a_{N}\}$ sampled from the search space. The learning of path filter is thus cast as a Positive and Unlabeled (PU) learning problem, where a binary predictor $\Phi$ is learned based on $\P$ and $\U$ so that the class labels of unseen architectures can be accurately predicted.

As an introduction of PU learning, we first investigate the expected risk (classification loss) on the whole dataset of the commonly supervised learning (PN learning) as 
\begin{equation} \label{eq:cls_risk}
    R(\Phi) = \pi_\P\mathbb{E}_\P[l_{+}(\Phi(\a))] + (1-\pi_\P)\mathbb{E}_\N[l_{-}(\Phi(\a))] ,
\end{equation}
where $\pi_\P = \mathbb{P}(y=+1)$ denotes the \textit{class prior} of positive data, $\N$ refers to negative dataset, and $l_+$ and $l_-$ denote classification losses with
\begin{align}
    \begin{split}
        \mathbb{E}_{\mathcal{P}}[l_+(\Phi(\bm{a}))] &= \frac1{|\P|} \sum_{\a\in\P} l(\Phi(\bm{a}),+1) ,\\
        \mathbb{E}_{\mathcal{N}}[l_-(\Phi(\bm{a}))] &= \frac1{|\N|} \sum_{\a\in\N} l(\Phi(\bm{a}),-1),
    \end{split}
\end{align}
which are the expectations of $l_+(\Phi(\bm{a}))$ on the positive dataset $\mathcal{P}$ and $l_-(\Phi(\bm{a}))$ on the negative dataset $\mathcal{N}$.

Nevertheless, the negative dataset $\N$ is unavailable in our PU learning setting. To train the model with positive and unlabeled data, the classical method uPU~\cite{du2015convex} encourages an unbiased formulation to the PN learning by rewriting the expectation of negative classification loss $\mathbb{E}_{\mathcal{N}}[l_-(\Phi(\bm{a}))]$ to 
%, we can rewrite the risk over the negative class-conditional distribution, \ie, 
\begin{equation} \label{eq:unbiased_negative}
    (1-\pi_\P)\mathbb{E}_\N[l_{-}(\Phi(\a))] = \mathbb{E}_\U[l_{-}(\Phi(\a))] - \pi_\P\mathbb{E}_\P[l_{-}(\Phi(\a))] ,
\end{equation}
%Based on Eq.(\ref{eq:unbiased_negative}), the classical method uPU~\cite{du2015convex} proposes an effective solution to PU learning by introducing the unbiased risk estimator
and thus Eq.(\ref{eq:cls_risk}) can be adapted to
\begin{equation}
    R(\Phi) = \pi_\P\mathbb{E}_\P[l_{+}(\Phi(\a))-l_{-}(\Phi(\a))] + \mathbb{E}_\U[l_{-}(\Phi(\a))] ,
\end{equation}
%which can achieve an unbiased estimation of the expected classification risk in Eq.(\ref{eq:cls_risk}). 

However, such a method easily leads to severe over-fitting, especially on deep neural classifiers. In our paper, to alleviate the above weakness, we leverage the learning objective in VPU~\cite{chen2020vpu}, which proposes a variational loss to approximate the ideal classifier through an upper-bound of Eq.(\ref{eq:cls_risk}), \ie, 
\begin{align}
    \label{eq:vpu}
    \begin{split}
        R(\Phi) &= \mbox{log}\mathbb{E}_\U[\Phi(\a)] - \mathbb{E}_{\P}[\mbox{log}\Phi(\a)] \\
        %&= \mbox{log}\frac{\sum_{\a\in\mathcal{B}_\U}\Phi(\a)}{B} - \frac{\sum_{\a\in\mathcal{B}_\P}\mbox{log}\Phi(\a)}{B} ,
    \end{split}
\end{align}
where mini-batches $\mathcal{B}_\U$ and $\mathcal{B}_\P$ are sampled from $\U$ and $\P$ with size $B$. By minimizing Eq.(\ref{eq:vpu}), we can obtain an effective binary classifier to distinguish good and weak paths.

\textbf{Train path filter with multi-path sampling.} We first construct a neural network as our path filter (binary classifier), which will be trained using PU learning. For the network structure, we follow a simple \textit{Embedding-RNN} pipeline as the previous work~\cite{zoph2016neural}. Concretely, as illustrated in Figure~\ref{fig:figure2} left, we use randomly initialized embeddings $\bm{E}\in \mathbb{R}^{L\times N \times H}$ to represent operations in the search space, where $L$, $N$, and $H$ are numbers of layers, candidate operations, and hidden dimensions, respectively, hence each operation in each layer is associated with an independent embedding. For example, the embedding of $j$-th operation $o_{j}^{(i)}$ in layer $i$ can be represented as $\bm{E}_{i,j}$ . For an input architecture $\bm{a} = (o^{(1)}, o^{(2)}, ..., o^{(L)})$, the predictor first encodes it through operation embedding $\bm{E}$ to get the hidden states $\bm{A}$ of the selected operations, where $\bm{A} \in \mathbb{R}^{L\times H}$. Then we use a bi-directional LSTM to get the feature $\bm{f}_a \in \mathbb{R}^{H}$ of the architecture. Finally, the architecture feature $\bm{f}_a$ is fed into a binary classifier (two-layer perceptions with intermediate ReLU activation) to obtain the prediction.

Following the multi-path sampling strategy in GreedyNAS, each time we train the path filter, we randomly sample $m$ paths and evaluate them using the loss on a small validation set, which contains $1000$ images sampled from the validation set. We sort those sampled paths with their losses in ascending order, and label the last $p$ percentage of paths as ``weak paths'' to build the positive dataset $\P$, while the unlabeled dataset $\U$ is constructed by uniformly sampling $10\times p\times m$ paths from the search space. With the learning objective in Eq.(\ref{eq:vpu}), we train the path filter with datasets $\P$ and $\U$ every $t$ epochs in the training of supernet to ensure its accuracy. Once the path filter is trained, it can be used to predict a batch of uniformly-sampled paths, and filter those paths with positive labels, and the remained paths are treated as the potentially-good paths and used in optimization.

\textbf{Stopping principle via path predictions.} In training, if the supernet is trained well, the rankings of paths tend to be steady; hence GreedyNAS proposes an early stopping principle by measuring the \textit{steadiness} of the candidate pool. We now propose a more accurate way by predicting more paths using the learned path filters, \ie, 
\begin{equation}
    u := \frac{\sum_{\bm{a}_i\in\A_r}\1_{\Phi_t(\bm{a}_i)=\Phi_{t-1}(\bm{a}_i)}}{M} > \beta,
\end{equation}
where $\A_r$ is a set of $M$ randomly sampled paths, $\Phi_t$ denotes the learned path filter at iteration $t$. $u$ measures the proportion of the same predictions in the last two path filters, if $u>\beta$, we believe the supernet has been trained enough, and its training can be stopped accordingly. We set $N = 10^4$ and $\beta = 0.9$ in our experiments.

\textbf{Evolutionary search with path filter.} We adopt evolutionary algorithm (EA) NSGA-II~\cite{deb2002fast} to search architectures. Unlike SPOS~\cite{guo2020single} generates architectures randomly and GreedyNAS~\cite{you2020greedynas} only specifies an initial population, we use the learned path filter to filter the weak architectures generated by EA during the whole search, thus the search could be more efficient.

\begin{algorithm}[t]
	\caption{Training supernet with a greedy path filter.}
	\label{alg:train_supernet}
	
	\begin{algorithmic}[1]
	    \REQUIRE{Supernet $\N$, path filter $\P$, max training iteration $N$, train dataset $\D_{tr}$, small validation dataset $\D_{val}$, predictor update interval $t$, number of evaluation paths $m$, weak path prior $q$, merge operation threshold $\eta$.}
	    %\ENSURE{x}
	    \FOR{$i=1,...,N$}
	        \STATE $\bm{a}_i\sim U(\N)$ ;\hfill$\triangleright$ \gray{sample one path uniformly}
	        \WHILE{is\_weak\_arch($\P$, $\bm{a}_i)$}
	            \STATE $\bm{a}_i\sim U(\N)$ ;
	        \ENDWHILE
	        \STATE train($\N$, $\bm{a}_i$, $\D_{tr}$); \hfill$\triangleright$ \gray{train for one iteration}
	        \IF{$i$ \% $t$ = 0}
	            \STATE sample $m$ paths $\A=\{\bm{a}_j\}_{j=1}^m$ i.i.d \wrt $\bm{a}_j\sim U(\N)$ ;
	            \STATE $\bm{s}=\{\mbox{evaluate}(\N, \bm{a}_j, \D_{val})\}^m_{j=1}$ ;
	            \STATE $\A_{weak} = \mbox{last}(\bm{s}, q)$; \hfill$\triangleright$ \gray{get last $q$ percentile paths}
	            \STATE train predictor $\P$ with $\A_{weak}$ ;
	            \STATE merge operations according to Section~\ref{sec:op_shrinkage};
	        \ENDIF
	    \ENDFOR
	    %\RETURN x
	\end{algorithmic}
\end{algorithm}
\begin{table*}[t]
	\renewcommand\arraystretch{1.3}
	\setlength\tabcolsep{1.6mm}
	\centering
	%\begin{center}
	\caption{Summary of our search spaces. Details can be found in Supplementary Materials.}
	\vspace{-2mm}
	\label{tab:search_space}
	\footnotesize
	%\normalsize
	\begin{tabular}{l|c|c|c|l}
		\Xhline{2\arrayrulewidth}
		Search space & Size & \#Layers & \#Operations & Operations \\
		\hline
		MB-SE & $13^{21}\approx2\times10^{23}$ & 21 & 13 & \{ MB3, MB6 \} $\times$ \{ K3, K5, K7, K3\_SE, K5\_SE, K7\_SE \} + \{ ID \} \\
		%\hline
		MB-SE+MixConv & $17^{21}\approx7\times10^{25}$ & 21 & 17 & MB-SE $\cup$ \{ MB3\_MIX, MB6\_MIX, MB3\_MIX\_SE, MB6\_MIX\_SE \}\\
		%\hline
		MB-SE+MixConv+Shuffle & $21^{21}\approx6\times10^{27}$ & 21 & 21 & MB-SE+MixConv $\cup$ \{ Shuffle\_3, Shuffle\_5, Shuffle\_7, Shuffle\_x \}\\
		\hline
		Res-50-SE & $19^{16}\approx3\times10^{20}$ & 16 & 19 & \{ ResNet, ResNeXt \} $\times$ \{ K3, K5, K7 \} $\times$ \{ 0.5$\times$, 1$\times$, 1.5$\times$ \} + \{ ID \} \\
		\Xhline{2\arrayrulewidth}
	\end{tabular}
	\vspace{-2mm}
	%\end{center}
\end{table*}

\subsection{Shrinking operations with learned embeddings} \label{sec:op_shrinkage}
The PU predictor distinguishes whether a path is good or bad using learned operation embeddings. If the embeddings of two operations $o_a^{(i)}$ and $o_b^{(i)}$ in layer $i$ are totally the same, it means that for all the architectures, replacing $o_a^{(i)}$ by $o_b^{(i)}$ would not affect the classification results and vice versa. As a result, if two operations act similarly, we can greedily merge them and keep the less-costly one (\eg, the one with smaller FLOPs).

Cosine similarity is a commonly-used metric to measure the similarity of two vectors. Given two vectors $\bm{x}$ and $\bm{y}$, their cosine similarity $S_c(\bm{x}, \bm{y})$ is represented as 
\begin{equation}
    S_c(\bm{x}, \bm{y}) = \frac{\bm{x}\cdot\bm{y}}{\|\bm{x}\|\cdot \|\bm{y}\|} = \frac{\sum_{i=1}^n x_iy_i}{\sqrt{\sum_{i=1}^n x_i^2}\sqrt{\sum_{i=1}^n y_i^2}} .
\end{equation}

After each time we train the predictor, we will measure the cosine similarities between different operations in each layer. If the similarity between two operations is less than a pre-defined threshold $s_{\text{thrd}}$, we then merge these two operations into one operation by keeping the one with smaller FLOPs and removing another one. 

Formally, for operations $\{o_1^{(i)}, o_2^{(i)}, ..., o_N^{(i)}\}$ in layer $i$, it has $\mathbb{C}_N^2$ combinations of pairs, we compute their cosine similarity between $o_j^{(i)}$ and $o_k^{(i)}$ using the learned embeddings, \ie,
\begin{equation}
    S_{j,k}^{(i)} = S_c(\bm{E}_{i,j}, \bm{E}_{i, k}) .
\end{equation}
For any layer $i\le L$ and operation pairs $o_j^{(i)}$ and $o_k^{(i)}$ ($j<k\le N$), we merge them into the one with less FLOPs when they satisfy $S_{j,k}^{(i)} > s_{\text{thrd}}$. After merging, the removed operations would never be sampled in training and search, thus reducing the training parameters in supernet. We set $s_{\text{thrd}} = 0.8$ in our experiments.

Our operation shrinkage method can significantly reduce the search space without additional evaluation steps. It can be naturally combined with the path-level shrinkage to perform a greedier search. The overall supernet training strategy is summarized in Algorithm~\ref{alg:train_supernet}.

\section{Experiments}
\subsection{Experimental setup}

\begin{table}[t]
	\renewcommand\arraystretch{1.1}
	\setlength\tabcolsep{1.17mm}
	\centering
	%\begin{center}
	\caption{Comparisons with our baseline methods on different scales of search spaces. Search spaces \textit{small}, \textit{medium}, and \textit{large} represent \textit{MB-SE}, \textit{MB-SE+MixConv}, and \textit{MB-SE+MixConv+Shuffle} in Table~\ref{tab:search_space}, respectively.}
	\vspace{-2mm}
	\label{tab:nas_comparison}
	\footnotesize
	%\normalsize
	\begin{tabular}{c|ccc|ccc}
		\Xhline{2\arrayrulewidth}
		\multirow{2}*{Method} & \multicolumn{3}{c|}{ACC (\%)} & \multicolumn{3}{c}{ACC on supernet (\%)}\\
		\cline{2-7}
		~ & small & medium & large & small & medium & large\\
		\hline
		SPOS~\cite{guo2020single} & 76.8 & 76.6 & 75.5 & 56.5 & 48.2 & 33.4\\
		GreedyNAS~\cite{you2020greedynas} & 77.1 & 76.8 & 76.5 & 57.6 & 49.3 & 35.1\\
		GreedyNASv2 & \textbf{77.3} & \textbf{77.4} & \textbf{77.5} & \textbf{58.1} & \textbf{55.5} & \textbf{43.8}\\
		\Xhline{2\arrayrulewidth}
	\end{tabular}
	\vspace{-2mm}
	%\end{center}
\end{table}
\begin{table*}[t]
	\renewcommand\arraystretch{1.17}
	\setlength\tabcolsep{4.5mm}
	\centering
	%\begin{center}
	\caption{Comparisons of searched architectures with state-of-the-art NAS methods and handcraft models. Training epochs and search number are hyper-parameters in the training of supernet. We measure the training cost of supernet using 8 NVIDIA V100 GPUs. *: trained with the same strategy as \textit{GreedyNASv2-L}.}
	\vspace{-2mm}
	\label{tab:imagenet}
	\footnotesize
	%\normalsize
	\begin{tabular}{r|cccc|ccc}
		\Xhline{2\arrayrulewidth}
		\multirow{2}*{Methods} & Top-1 & Top-5 & FLOPs & Params & Training & Training cost & Search\\
		%\cline{6-8}
		~ & (\%) & (\%) & (M) & (M) & epochs & (GPU days) & number\\
		\hline
		\multicolumn{8}{c}{mobile search space}\\
		\hline
		MobileNetV2~\cite{sandler2018mobilenetv2} & 72.0 & 91.0 & 300 & 3.4 & - & - & -\\
		%MnasNet-A1 & 75.2 & 92.5 & 312 & 3.9 & - & 288 & -\\
		EfficientNet-B0~\cite{tan2019efficientnet} & 76.3 & 93.2 & 390 & 5.3 & - & - & - \\
		%ABS~\cite{hu2020angle} & 74.4 & - & 325 & - & 105 & - & -\\
		SPOS~\cite{guo2020single} & 74.7 & - & 328 & 3.4 & 120 & 12 & 1000 \\ 
		%BS-NAS-A~\cite{shen2020bs} & 75.9 & 92.8 & 465 & 4.9 & 240 & 10 & 1500 \\
		MCT-NAS-B~\cite{su2021prioritized} & 76.9 & 93.4 & 327 & 6.3 & 120 & 12 & 100 \\
		K-shot-NAS-B~\cite{su2021k} & 77.2 & 93.3 & 332 & 6.2 & 120 & 12 & 1000 \\
		NSENet~\cite{ci2021evolving} & 77.3 & - & 333 & 7.6 & 100 & 166.7 & 2100 \\
		GreedyNAS-B~\cite{you2020greedynas} & 76.8 & 93.0 & 324 & 5.2 & 46 & 7 & 1000 \\
		\textbf{GreedyNASv2-S} & \textbf{77.5} & 93.5 & 324 & 5.7 & 65 & 7 & 500\\
		\hline
		\multicolumn{8}{c}{ResNet search space}\\
		\hline
		ResNeXt-50~\cite{xie2017aggregated} & 77.8 & - & 4230 & 25.0 & - & - & -\\
		RegNetX-4.0GF~\cite{radosavovic2020designing} & 78.6 & - & 3964 & 22.1 & - & - & -\\
		ResNet-50$^*$~\cite{he2016deep} & 78.8 & 94.6 & 4089 & 25.6 & - & - & -\\
		SE-ResNeXt-50~\cite{hu2018squeeze} & 78.9 & 94.5 & 4233 & 27.6 & - & - & -\\
		SKNet-50~\cite{li2019selective} & 79.2 & - & 4470 & 27.5 & - & - & -\\
		SE-ResNet-50$^*$~\cite{hu2018squeeze} & 80.5 & 94.8 & 4094 & 30.6 & - & -\\
		\textbf{GreedyNASv2-L} & \textbf{81.1} & 95.4 & 4098 & 26.9 & 57 & 9 & 500\\
		\Xhline{2\arrayrulewidth}
	\end{tabular}
	\vspace{-2mm}
	%\end{center}
\end{table*}

\textbf{Search space.} As summarized in Table~\ref{tab:search_space}, for comparisons with baselines~\cite{you2020greedynas, su2021prioritized}, we first search on MobileNetV2-SE search space, which consists of Identity, MobileNetV2 block~\cite{sandler2018mobilenetv2}, and optional SE modules~\cite{hu2018squeeze}. To validate our superiority on larger search spaces, we extend the search space with MixConv~\cite{tan2019mixconv} block, namely, \textit{MobileNetV2-SE+MixConv}. Moreover, we set up an extremely-large search space~\textit{MobileNetV2-SE+MixConv+Shuffle}, which adds $4$ ShuffleNetV2 blocks~\cite{ma2018shufflenet} following SPOS~\cite{guo2020single}. %\blue{Since the computation of ResNet like structures is more efficient than mobile networks with depth-wise convolution on GPUs}, 
In order to validate the effectiveness of our method on larger networks, we also introduce a ResNet-like search space, which consists of the blocks in ResNet~\cite{he2016deep}, ResNeXt~\cite{xie2017aggregated}, and SENet~\cite{hu2018squeeze}. Details can be found in supplementary materials.

\textbf{Supernet.} We randomly sampled $50$k images from ILSVRC-2012~\cite{imagenet} train set to build our validation set, and the remains are used as the training set. We train the supernet with an SGD optimizer and a total batch size of $1024$, a cosine learning rate which decays $120$ epochs with an initial learning rate $0.12$ is adopted. In the first $20$ epochs of training, we sample architectures uniformly for warm-up, then train the path filter every $5$ epoch and use it to sample architectures. The weak path prior $q$ is increased from $0.5$ to $0.99$ in $90$ epochs.

\textbf{Path filter.} The path filter is constructed by an embedding with $128$ dimensions, followed by a bidirectional LSTM and two fully-connected layers with intermediate ReLU activation, all the hidden dimensions are set to $128$. We train the path filter for $3000$ iterations after every $5$ epoch of the supernet training, an Adam optimizer with batch size $1024$ and weight decay $5\times10^{-3}$ is adopted, the learning rate is set to $10^{-3}$.

\textbf{Search.} We use the learned path filter to help the evolutionary algorithm NSGA-II~\cite{deb2002fast} search architectures. The search number is set to $500$. 

\textbf{Retraining.} In retraining, we use the official ILSVRC-2012~\cite{imagenet} train set and report the accuracy on the original validation set. Following \cite{you2020greedynas, su2021prioritized}, we train the searched mobile architectures using a RMSProp optimizer with a batch size $96$ on each of $8$ GPU, a step learning rate scheduler which warmups for $3$ epochs then decays $0.97$ every $2.4$ epochs is adopted with initial value $0.048$. While for ResNet-like model, we train it using SGD optimizer with weight decay $10^{-4}$ and batch size $1536$, the initial learning rate is set to $0.6$ and decays for $240$ epochs with a cosine scheduler.
We use a data augmentation pipeline of Autoaugment~\cite{cubuk2019autoaugment}, random cropping, and clipping. We use a train and test image size of $224\times224$. Besides, an exponential moving average on weights is also adopted with a decay $0.9999$.

\subsection{Results on ImageNet}
\textbf{Comparisons with NAS methods.} We first compare our GreedyNASv2 with the baseline methods SPOS~\cite{guo2020single} and GreedyNAS~\cite{you2020greedynas} on \textit{MB-SE}, \textit{MB-SE+MixConv}, and \textit{MB-SE+MixConv+Shuffle} search spaces based on our implementations. We use a constraint of $330$M FLOPs and report the evaluation accuracies of the searched architectures on retraining and supernet in Table~\ref{tab:nas_comparison}. We can see that, on all sizes of search spaces, our GreedyNASv2 can obtain higher ACCs than the other two methods. Moreover, the performance of SPOS on medium and large spaces drops significantly, showing that it would be difficult for SPOS to train promising supernets on such huge spaces. While our GreedyNASv2 obtains similar performance, and even achieves the best performance on large space. We compare our obtained model \textit{GreedyNASv2-S} on \textit{MB-SE+MixConv+Shuffle} search space with state-of-the-art NAS methods in Table~\ref{tab:imagenet}.

\textbf{Search for larger networks.} To evaluate our generalization, we conduct search on a ResNet-style search space \textit{Res-50-SE}. As the results shown in Table~\ref{tab:imagenet}, our GreedyNASv2 achieves significant improvement compared to the baseline ResNet, ResNeXt, and SENet models. Note that we train our \textit{GreedyNASv2-L} with simple SGD optimizer and an additional Autoaugment~\cite{cubuk2019autoaugment} data pipeline. However, its performance still outperforms the state-of-the-art training strategies with more sophisticated optimization and strong data augmentation in TIMM ~\cite{wightman2021resnet}, which achieves 80.4\% ACC on ResNet-50.

\subsection{Results on NAS-Bench-Macro}
MCT-NAS~\cite{su2021prioritized} proposes a NAS benchmark named NAS-Bench-Macro for single path one-shot NAS methods, which consists of $6561$ architectures and their isolated evaluation results on CIFAR-10 dataset. We leverage this benchmark to validate the effectiveness of GreedyNASv2.

\textbf{Performance of path filter with ground-truth training data.} To validate the pure performance of our PU learning method, we conduct experiments to train the path filter with the ground-truth labels in the benchmark. Concretely, we split the architectures to $10\%$ of good paths and $90\%$ of weak paths according to their evaluation accuracies, then samples $1\%,10\%,50\%$, and $100\%$ data as a train set. We use the whole set to validate the classification performance of the learned path filter. We use precision and recall as evaluation metrics. We first train the path filter with our PU learning settings using only weak paths and randomly sampled unlabeled paths. For comparisons, we also adopt the PN learning (traditional supervised learning) by using both weak labels and good labels. As the results reported in Table~\ref{tab:nasbench}, the PU learning even achieves better performance than PN learning. This might be because paths are densely distributed in the space, an absolute threshold for partitioning ``P'' and ``N'' data might involve many uncertain paths, while the PU learning could handle this uncertainty well by treating unlabeled data more safely.

\begin{table}[t]
	\renewcommand\arraystretch{1.1}
	\setlength\tabcolsep{1.mm}
	\centering
	%\begin{center}
	\caption{Performance of our PU learning method compared with PN learning on NAS-Bench-Macro~\cite{su2021prioritized}.}
	\vspace{-2mm}
	\label{tab:nasbench}
	\footnotesize
	%\normalsize
	\begin{tabular}{c|cc|cc|cc|cc}
		\Xhline{2\arrayrulewidth}
		\multirow{2}*{Method} & \multicolumn{2}{c|}{1\% data} & \multicolumn{2}{c|}{10\% data} & \multicolumn{2}{c|}{50\% data} & \multicolumn{2}{c}{100\% data}\\
		\cline{2-9}
		~ & Pre. & Recall & Pre. & Recall & Pre. & Recall & Pre. & Recall\\
		\hline
		PU & 97.21 & 97.08 & 98.37 & 98.34 & 98.30 & 98.19 & 98.81 & 98.62\\
		PN & 79.52 & 85.82 & 78.74 & 93.14 & 88.45 & 85.21 & 85.55 & 90.24\\
		\Xhline{2\arrayrulewidth}
	\end{tabular}
	%\end{center}
\end{table}

\begin{figure}[t]
    \centering
    \includegraphics[width=0.75\linewidth]{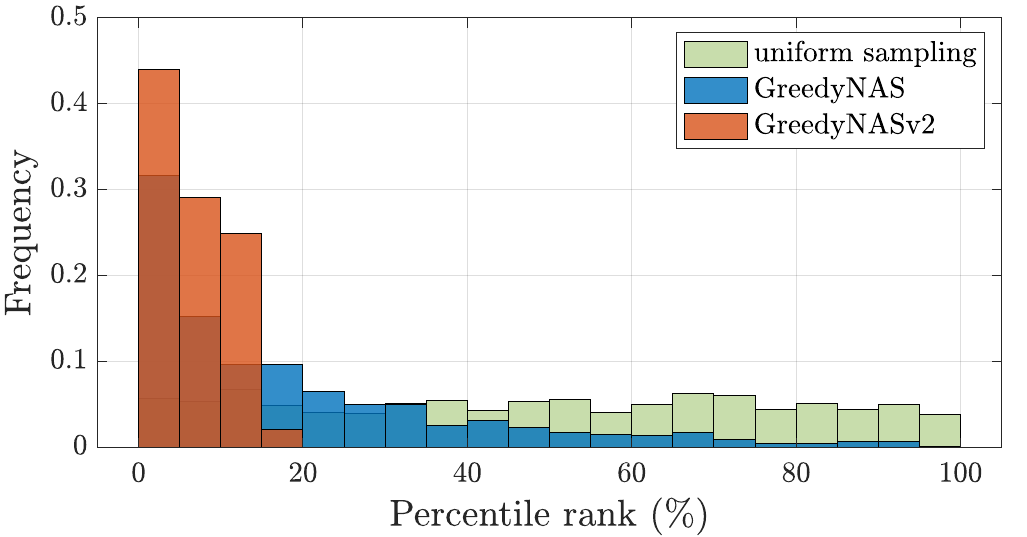}
    \vspace{-2mm}
    \caption{Histogram of percentile rank of sampled paths on NAS-Bench-Macro search space. The average percentile rank of \textit{uniform sampling}, \textit{GreedyNAS}, and \textit{GreedyNASv2} are $50.6\%$, $18.1\%$, and $6.4\%$, respectively.}
    \label{fig:percentile_rank}
    \vspace{-4mm}
\end{figure}

\textbf{Performance of path filter on supernet.} We also validate the performance of the path filter learned in supernet training. Unlike the previous experiment using the ground-truth labels, the labels in supernet training are generated by evaluating sampled architectures with a validation set; hence, they could have some noises. The learned path filter obtains $93.74\%$ precision and $98.21\%$ recall on the whole search space, comparing to the best performance of using ground-truth labels ($98.81\%$ precision and $98.62\%$ recall), the small decrease in precision is acceptable since our method only needs to greedily focus on a proportion of potentially-good paths instead of locating all the good ones.

\textbf{Average percentile ranks of the sampled architectures during training.} We collect the percentile ranks of the sampled architectures during training. As shown in Figure~\ref{fig:percentile_rank}, our method samples more paths with smaller percentile ranks compared to baselines, which means that our trained supernet is greedier towards those good paths.

\textbf{Correlation between validation and retraining accuracies.} Since GreedyNASv2 greedily filters out weak paths and focuses on the potential good paths, the correlation between validation accuracies on supernet and their retraining accuracies will be boosted in terms of those good paths identified by the path filter. We measure the rank correlations on those paths within top 10\% percentile ranks on NAS-Bench-Macro, and the Kendall's Tau of SPOS, GreedyNAS, and GreedyNASv2 are 23.9\%, 41.5\%, and 50.3\%, respectively. This indicates our effectiveness since discriminating among good paths are fairly challenging. 

\subsection{Ablation studies}

\begin{figure}
    \centering
    \includegraphics[width=0.9\linewidth]{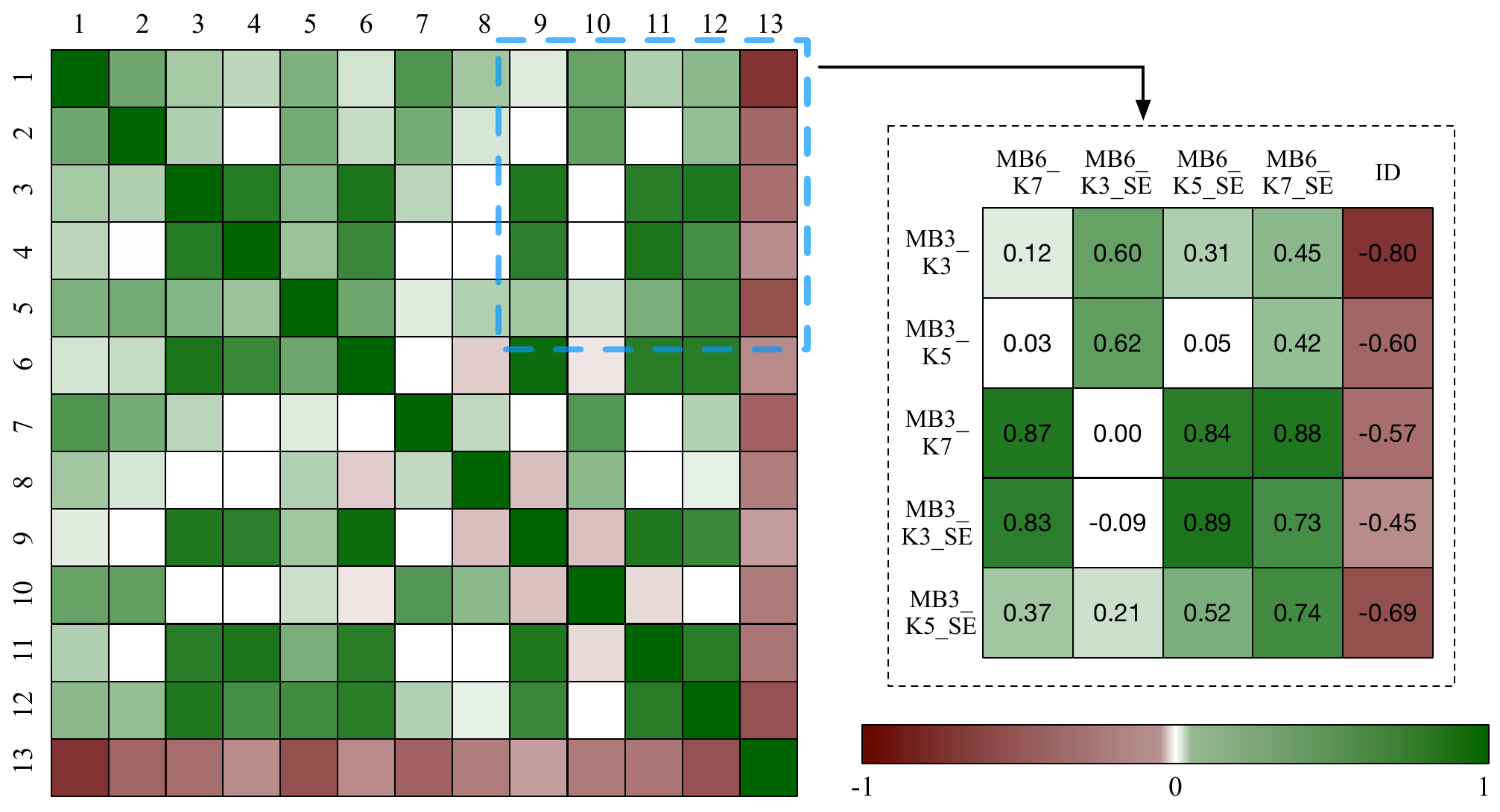}
    \makeatletter\def\@captype{figure}\makeatother
    \vspace{-2mm}
    \caption{Visualization of learned operation similarities at the 1st searching layer of MB-SE supernet.}
    \label{fig:sim}
    
    \centering
    \setlength\tabcolsep{2.2mm}
    \renewcommand\arraystretch{1.1}
    \footnotesize
    \makeatletter\def\@captype{table}\makeatother
    \caption{Summary of merged operations (partial) \wrt Figure~\ref{fig:sim}.}
    \begin{tabular}{ll|c|l}
		\Xhline{2\arrayrulewidth}
		\multicolumn{2}{c|}{Operation pair} & Similarity & Retained operation \\
		\hline
		MB3\_K7\_SE & MB6\_K7 & 0.95 & MB3\_K7\_SE\\
		MB3\_K7 & MB3\_K7\_SE & 0.89 & MB3\_K7\\
		MB3\_K3\_SE & MB6\_K5\_SE & 0.89 & MB3\_K3\_SE\\
		MB3\_K7 & MB6\_K7\_SE & 0.88 & MB3\_K7\\
		MB3\_K7 & MB6\_K7 & 0.87 & MB3\_K7\\
		\Xhline{2\arrayrulewidth}
	\end{tabular}
	\vspace{-2mm}
    \label{tab:merge_ops}
\end{figure}

\textbf{Visualization of learned operation similarities.} As summarized in Figure~\ref{fig:sim} and Table~\ref{tab:merge_ops}, we visualize the learned operation similarities at the first searching layer of MB-SE supernet. It shows that the operations with the same kernel size are more likely to have similar embeddings. For \textit{ID} operation, it has negative similarities to all the other operations since it performs poorly on down-sampling layers. For the whole supernet, there are a total of $35$ ($\sim 13\%$) out of $13\times 21 = 273$ operations merged.

\section{Conclusion}
We propose GreedyNASv2, a NAS method with greedy path-level and operation-level shrinkage of search space. Unlike the previous works, our method achieves a greedier search with a greedy path filter, which is trained with highly-confident ``weak'' paths and unlabeled paths using positive-unlabeled (PU) learning. By dint of the learned embeddings in our path filter, we can further perform operation-level shrinkage by aggregating similar operations with similar embeddings, and the search can be more efficient and accurate. Extensive experiments show that our GreedyNASv2 achieves better performance compared to our baselines in various scales of search spaces. 

\section*{Acknowledgements}
This work was supported in part by the Australian Research Council under Project DP210101859 and the University of Sydney SOAR Prize.

%%%%%%%%% REFERENCES
{\small
\bibliographystyle{ieee_fullname}
\bibliography{main}

\begin{thebibliography}{10}\itemsep=-1pt

\bibitem{chen2020vpu}
Hui Chen, Fangqing Liu, Yin Wang, Liyue Zhao, and Hao Wu.
\newblock A variational approach for learning from positive and unlabeled data.
\newblock In H. Larochelle, M. Ranzato, R. Hadsell, M.~F. Balcan, and H. Lin,
  editors, {\em Advances in Neural Information Processing Systems}, volume~33,
  pages 14844--14854. Curran Associates, Inc., 2020.

\bibitem{mmdetection}
Kai Chen, Jiaqi Wang, Jiangmiao Pang, Yuhang Cao, Yu Xiong, Xiaoxiao Li,
  Shuyang Sun, Wansen Feng, Ziwei Liu, Jiarui Xu, Zheng Zhang, Dazhi Cheng,
  Chenchen Zhu, Tianheng Cheng, Qijie Zhao, Buyu Li, Xin Lu, Rui Zhu, Yue Wu,
  Jifeng Dai, Jingdong Wang, Jianping Shi, Wanli Ouyang, Chen~Change Loy, and
  Dahua Lin.
\newblock {MMDetection}: Open mmlab detection toolbox and benchmark.
\newblock {\em arXiv preprint arXiv:1906.07155}, 2019.

\bibitem{chen2019detnas}
Yukang Chen, Tong Yang, Xiangyu Zhang, Gaofeng Meng, Xinyu Xiao, and Jian Sun.
\newblock Detnas: Backbone search for object detection.
\newblock {\em Advances in Neural Information Processing Systems},
  32:6642--6652, 2019.

\bibitem{chu2021fairnas}
Xiangxiang Chu, Bo Zhang, and Ruijun Xu.
\newblock Fairnas: Rethinking evaluation fairness of weight sharing neural
  architecture search.
\newblock In {\em Proceedings of the IEEE/CVF International Conference on
  Computer Vision}, pages 12239--12248, 2021.

\bibitem{ci2021evolving}
Yuanzheng Ci, Chen Lin, Ming Sun, Boyu Chen, Hongwen Zhang, and Wanli Ouyang.
\newblock Evolving search space for neural architecture search.
\newblock In {\em Proceedings of the IEEE/CVF International Conference on
  Computer Vision}, pages 6659--6669, 2021.

\bibitem{cubuk2019autoaugment}
Ekin~D Cubuk, Barret Zoph, Dandelion Mane, Vijay Vasudevan, and Quoc~V Le.
\newblock Autoaugment: Learning augmentation strategies from data.
\newblock In {\em Proceedings of the IEEE/CVF Conference on Computer Vision and
  Pattern Recognition}, pages 113--123, 2019.

\bibitem{deb2002fast}
Kalyanmoy Deb, Amrit Pratap, Sameer Agarwal, and TAMT Meyarivan.
\newblock A fast and elitist multiobjective genetic algorithm: Nsga-ii.
\newblock {\em IEEE transactions on evolutionary computation}, 6(2):182--197,
  2002.

\bibitem{imagenet}
Jia Deng, Wei Dong, Richard Socher, Li-Jia Li, Kai Li, and Li Fei-Fei.
\newblock Imagenet: A large-scale hierarchical image database.
\newblock In {\em 2009 IEEE conference on computer vision and pattern
  recognition}, pages 248--255. Ieee, 2009.

\bibitem{dodge2008concise}
Yadolah Dodge.
\newblock {\em The concise encyclopedia of statistics}.
\newblock Springer Science \& Business Media, 2008.

\bibitem{du2015convex}
Marthinus Du~Plessis, Gang Niu, and Masashi Sugiyama.
\newblock Convex formulation for learning from positive and unlabeled data.
\newblock In {\em International conference on machine learning}, pages
  1386--1394. PMLR, 2015.

\bibitem{elkan2008learning}
Charles Elkan and Keith Noto.
\newblock Learning classifiers from only positive and unlabeled data.
\newblock In {\em Proceedings of the 14th ACM SIGKDD international conference
  on Knowledge discovery and data mining}, pages 213--220, 2008.

\bibitem{guo2020hit}
Jianyuan Guo, Kai Han, Yunhe Wang, Chao Zhang, Zhaohui Yang, Han Wu, Xinghao
  Chen, and Chang Xu.
\newblock Hit-detector: Hierarchical trinity architecture search for object
  detection.
\newblock In {\em Proceedings of the IEEE/CVF Conference on Computer Vision and
  Pattern Recognition}, pages 11405--11414, 2020.

\bibitem{guo2020single}
Zichao Guo, Xiangyu Zhang, Haoyuan Mu, Wen Heng, Zechun Liu, Yichen Wei, and
  Jian Sun.
\newblock Single path one-shot neural architecture search with uniform
  sampling.
\newblock In {\em European Conference on Computer Vision}, pages 544--560.
  Springer, 2020.

\bibitem{he2016deep}
Kaiming He, Xiangyu Zhang, Shaoqing Ren, and Jian Sun.
\newblock Deep residual learning for image recognition.
\newblock In {\em Proceedings of the IEEE conference on computer vision and
  pattern recognition}, pages 770--778, 2016.

\bibitem{hu2018squeeze}
Jie Hu, Li Shen, and Gang Sun.
\newblock Squeeze-and-excitation networks.
\newblock In {\em Proceedings of the IEEE conference on computer vision and
  pattern recognition}, pages 7132--7141, 2018.

\bibitem{hu2020angle}
Yiming Hu, Yuding Liang, Zichao Guo, Ruosi Wan, Xiangyu Zhang, Yichen Wei,
  Qingyi Gu, and Jian Sun.
\newblock Angle-based search space shrinking for neural architecture search.
\newblock In {\em European Conference on Computer Vision}, pages 119--134.
  Springer, 2020.

\bibitem{huang2020explicitly}
Tao Huang, Shan You, Yibo Yang, Zhuozhuo Tu, Fei Wang, Chen Qian, and Changshui
  Zhang.
\newblock Explicitly learning topology for differentiable neural architecture
  search.
\newblock {\em arXiv preprint arXiv:2011.09300}, 2020.

\bibitem{kendall1938new}
Maurice~G Kendall.
\newblock A new measure of rank correlation.
\newblock {\em Biometrika}, 30(1/2):81--93, 1938.

\bibitem{kiryo2017positive}
Ryuichi Kiryo, Gang Niu, Marthinus~C du Plessis, and Masashi Sugiyama.
\newblock Positive-unlabeled learning with non-negative risk estimator.
\newblock In {\em Proceedings of the 31st International Conference on Neural
  Information Processing Systems}, pages 1674--1684, 2017.

\bibitem{letouzey2000learning}
Fabien Letouzey, Fran{\c{c}}ois Denis, and R{\'e}mi Gilleron.
\newblock Learning from positive and unlabeled examples.
\newblock In {\em International Conference on Algorithmic Learning Theory},
  pages 71--85. Springer, 2000.

\bibitem{li2019selective}
Xiang Li, Wenhai Wang, Xiaolin Hu, and Jian Yang.
\newblock Selective kernel networks.
\newblock In {\em Proceedings of the IEEE/CVF Conference on Computer Vision and
  Pattern Recognition}, pages 510--519, 2019.

\bibitem{lin2017feature}
Tsung-Yi Lin, Piotr Doll{\'a}r, Ross Girshick, Kaiming He, Bharath Hariharan,
  and Serge Belongie.
\newblock Feature pyramid networks for object detection.
\newblock In {\em Proceedings of the IEEE conference on computer vision and
  pattern recognition}, pages 2117--2125, 2017.

\bibitem{lin2017focal}
Tsung-Yi Lin, Priya Goyal, Ross Girshick, Kaiming He, and Piotr Doll{\'a}r.
\newblock Focal loss for dense object detection.
\newblock In {\em Proceedings of the IEEE international conference on computer
  vision}, pages 2980--2988, 2017.

\bibitem{lin2014microsoft}
Tsung-Yi Lin, Michael Maire, Serge Belongie, James Hays, Pietro Perona, Deva
  Ramanan, Piotr Doll{\'a}r, and C~Lawrence Zitnick.
\newblock Microsoft coco: Common objects in context.
\newblock In {\em European conference on computer vision}, pages 740--755.
  Springer, 2014.

\bibitem{ma2018shufflenet}
Ningning Ma, Xiangyu Zhang, Hai-Tao Zheng, and Jian Sun.
\newblock Shufflenet v2: Practical guidelines for efficient cnn architecture
  design.
\newblock In {\em Proceedings of the European conference on computer vision
  (ECCV)}, pages 116--131, 2018.

\bibitem{radosavovic2020designing}
Ilija Radosavovic, Raj~Prateek Kosaraju, Ross Girshick, Kaiming He, and Piotr
  Doll{\'a}r.
\newblock Designing network design spaces.
\newblock In {\em Proceedings of the IEEE/CVF Conference on Computer Vision and
  Pattern Recognition}, pages 10428--10436, 2020.

\bibitem{sandler2018mobilenetv2}
Mark Sandler, Andrew Howard, Menglong Zhu, Andrey Zhmoginov, and Liang-Chieh
  Chen.
\newblock Mobilenetv2: Inverted residuals and linear bottlenecks.
\newblock In {\em Proceedings of the IEEE conference on computer vision and
  pattern recognition}, pages 4510--4520, 2018.

\bibitem{shen2020bs}
Zan Shen, Jiang Qian, Bojin Zhuang, Shaojun Wang, and Jing Xiao.
\newblock Bs-nas: Broadening-and-shrinking one-shot nas with searchable numbers
  of channels.
\newblock {\em arXiv preprint arXiv:2003.09821}, 2020.

\bibitem{su2021prioritized}
Xiu Su, Tao Huang, Yanxi Li, Shan You, Fei Wang, Chen Qian, Changshui Zhang,
  and Chang Xu.
\newblock Prioritized architecture sampling with monto-carlo tree search.
\newblock In {\em Proceedings of the IEEE/CVF Conference on Computer Vision and
  Pattern Recognition}, pages 10968--10977, 2021.

\bibitem{su2020locally}
Xiu Su, Shan You, Tao Huang, Fei Wang, Chen Qian, Changshui Zhang, and Chang
  Xu.
\newblock Locally free weight sharing for network width search.
\newblock In {\em International Conference on Learning Representations}, 2020.

\bibitem{su2021bcnet}
Xiu Su, Shan You, Fei Wang, Chen Qian, Changshui Zhang, and Chang Xu.
\newblock Bcnet: Searching for network width with bilaterally coupled network.
\newblock In {\em Proceedings of the IEEE/CVF Conference on Computer Vision and
  Pattern Recognition}, pages 2175--2184, 2021.

\bibitem{su2021vision}
Xiu Su, Shan You, Jiyang Xie, Mingkai Zheng, Fei Wang, Chen Qian, Changshui
  Zhang, Xiaogang Wang, and Chang Xu.
\newblock Vision transformer architecture search.
\newblock {\em arXiv preprint arXiv:2106.13700}, 2021.

\bibitem{su2021k}
Xiu Su, Shan You, Mingkai Zheng, Fei Wang, Chen Qian, Changshui Zhang, and
  Chang Xu.
\newblock K-shot {NAS:} learnable weight-sharing for {NAS} with k-shot
  supernets.
\newblock In {\em {ICML}}, volume 139 of {\em Proceedings of Machine Learning
  Research}, pages 9880--9890. {PMLR}, 2021.

\bibitem{tan2019efficientnet}
Mingxing Tan and Quoc Le.
\newblock Efficientnet: Rethinking model scaling for convolutional neural
  networks.
\newblock In {\em International Conference on Machine Learning}, pages
  6105--6114. PMLR, 2019.

\bibitem{tan2019mixconv}
Mingxing Tan and Quoc~V. Le.
\newblock Mixconv: Mixed depthwise convolutional kernels.
\newblock In {\em {BMVC}}, page~74. {BMVA} Press, 2019.

\bibitem{wightman2021resnet}
Ross Wightman, Hugo Touvron, and Herv{\'e} J{\'e}gou.
\newblock Resnet strikes back: An improved training procedure in timm.
\newblock {\em arXiv preprint arXiv:2110.00476}, 2021.

\bibitem{xie2017aggregated}
Saining Xie, Ross Girshick, Piotr Doll{\'a}r, Zhuowen Tu, and Kaiming He.
\newblock Aggregated residual transformations for deep neural networks.
\newblock In {\em Proceedings of the IEEE conference on computer vision and
  pattern recognition}, pages 1492--1500, 2017.

\bibitem{yang2020ista}
Yibo Yang, Hongyang Li, Shan You, Fei Wang, Chen Qian, and Zhouchen Lin.
\newblock Ista-nas: Efficient and consistent neural architecture search by
  sparse coding.
\newblock {\em arXiv preprint arXiv:2010.06176}, 2020.

\bibitem{yang2021towards}
Yibo Yang, Shan You, Hongyang Li, Fei Wang, Chen Qian, and Zhouchen Lin.
\newblock Towards improving the consistency, efficiency, and flexibility of
  differentiable neural architecture search.
\newblock In {\em Proceedings of the IEEE/CVF Conference on Computer Vision and
  Pattern Recognition}, pages 6667--6676, 2021.

\bibitem{you2020greedynas}
Shan You, Tao Huang, Mingmin Yang, Fei Wang, Chen Qian, and Changshui Zhang.
\newblock Greedynas: Towards fast one-shot nas with greedy supernet.
\newblock In {\em Proceedings of the IEEE/CVF Conference on Computer Vision and
  Pattern Recognition}, pages 1999--2008, 2020.

\bibitem{zhang2018mixup}
Hongyi Zhang, Moustapha Cisse, Yann~N Dauphin, and David Lopez-Paz.
\newblock mixup: Beyond empirical risk minimization.
\newblock In {\em International Conference on Learning Representations}, 2018.

\bibitem{zoph2016neural}
Barret Zoph and Quoc~V Le.
\newblock Neural architecture search with reinforcement learning.
\newblock {\em arXiv preprint arXiv:1611.01578}, 2016.

\end{thebibliography}
}

\newpage
\appendix
\onecolumn

\section{Details of Search Spaces}
\subsection{Macro structures}
In this paper, we conduct experiments on two macro structures of supernet, as presented in Table~\ref{tab:structure_mbv2} and Table~\ref{tab:structure_res50}. The MobileNetV2 supernet is used for \textit{MB-SE}, \textit{MB-SE+MixConv}, and \textit{MB-SE+MixConv+shuffle} search spaces, while the \textit{Res-50-SE} adopts the same structure as ResNet-50~\cite{he2016deep} in Table~\ref{tab:structure_res50}.

\begin{minipage}{\textwidth}
\vspace{6mm}
\begin{minipage}[c]{0.5\textwidth}
	\renewcommand\arraystretch{1.2}
	\setlength\tabcolsep{3mm}
	\centering
	%\begin{center}
	\makeatletter\def\@captype{table}\makeatother
	\caption{Macro structure of our MobileNetV2 search space. \textit{input} denotes the input feature size for each layer, \textit{channels} means the output channels of the layer, \textit{repeat} denotes the repeat times of stacking the same blocks, and \textit{stride} is for the stride of first block when stacked for multiple times.}
	\vspace{-2mm}
	\label{tab:structure_mbv2}
	\footnotesize
	%\normalsize
	\begin{tabular}{c|c|c|c|c}
		\Xhline{2\arrayrulewidth}
		input & block & channels & repeat & stride \\
		\hline
		$224^2\times3$ & $3\times3$ conv & 32 & 1 & 2\\
		$112^2\times 32$ & MB1\_K3 & 16 & 1 & 1 \\
		$112^2\times 16$ & Choice Block & 32 & 4 & 2 \\
		$56^2\times 32$ & Choice Block & 40 & 4 & 2 \\
		$28^2\times 40$ & Choice Block & 80 & 4 & 2 \\
		$14^2\times 80$ & Choice Block & 96 & 4 & 1 \\
		$14^2\times 96$ & Choice Block & 192 & 4 & 2 \\
		$7^2\times 192$ & Choice Block & 320 & 1 & 1 \\
		$7^2\times 320$ & $1\times 1$ conv & 1280 & 1 & 1 \\ 
		$7^2\times 1280$ & global avgpool & - & 1 & -\\ 
		$1280$ & FC & 1000 & 1& -\\ 
		\Xhline{2\arrayrulewidth}
	\end{tabular}
	%\end{center}
\end{minipage}
\hfill
\begin{minipage}[c]{0.5\textwidth}
	\renewcommand\arraystretch{1.2}
	\setlength\tabcolsep{2mm}
	\centering
	%\begin{center}
	\makeatletter\def\@captype{table}\makeatother
	\caption{Macro structure of our Res-50-SE search space.}
	\vspace{-2mm}
	\label{tab:structure_res50}
	\footnotesize
	%\normalsize
	\begin{tabular}{c|c|c|c|c}
		\Xhline{2\arrayrulewidth}
		input & block & channels & repeat & stride \\
		\hline
		$224^2\times3$ & $7\times7$ conv & 64 & 1 & 2\\
		$112^2\times 64$ & $3\times3$ max pool & 64 & 1 & 2 \\
		$56^2\times 64$ & Choice Block & 256 & 3 & 1 \\
		$56^2\times 256$ & Choice Block & 512 & 4 & 2 \\
		$28^2\times 512$ & Choice Block & 1024 & 6 & 2 \\
		$14^2\times 1024$ & Choice Block & 2048 & 3 & 2 \\
		$7^2\times 2048$ & global avg pool & - & 1 & -\\ 
		$2048$ & FC & 1000 & 1& -\\ 
		\Xhline{2\arrayrulewidth}
	\end{tabular}
	%\end{center}
\end{minipage}
\vspace{6mm}
\end{minipage}

\subsection{Candidate operations}
The candidate operations in \textit{Choice Block} of each supernet are summarized as follows.

$\bullet$ \textbf{MB-SE.} Following the previous NAS methods~\cite{you2020greedynas,su2021prioritized,su2021k}, we conduct the same searching operations in \textit{MB-SE} search space, which consists of $13$ MobileNetV2~\cite{sandler2018mobilenetv2} blocks with optional SE~\cite{hu2018squeeze} module, as summarized in Table~\ref{tab:op_mbv2}.

$\bullet$ \textbf{MB-SE+MixConv.} We design a new MobileNetV2 search space with additional MixConv blocks~\cite{tan2019mixconv}, which aims to mix the outputs of different kernel sizes ($3\times3$, $5\times5$, and $7\times7$) of depth-wise convolution in MobileNetV2 block.

$\bullet$ \textbf{MB-SE+MixConv+Shuffle.} To further validate our performance on a larger search space, we add the ShuffleNetV2~\cite{ma2018shufflenet} blocks in SPOS~\cite{guo2020single}.

$\bullet$ \textbf{Res-50-SE.} We leverage the blocks in ResNet~\cite{he2016deep} and ResNeXt~\cite{xie2017aggregated} to build our \textit{Res-50-SE} search space, and all of them are equipped with additional SE modules. We design blocks with different kernel sizes ($3$, $5$, and $7$), and a \textit{ratio} is used to control the intermediate number of channels, which has choices $0.5$, $1.0$, and $1.5$, \eg, $0.5$ means that the number of intermediate channels is $0.5\times$ compared to the original one. The total number of candidate operations is $19$, with an additional \textit{ID} operation for layer removal.

The detailed settings of candidate operations are summarized in Table~\ref{tab:op_mbv2} and Table~\ref{tab:op_res50}.

\begin{minipage}{\textwidth}
\vspace{6mm}
\begin{minipage}[c]{0.5\textwidth}
	\renewcommand\arraystretch{1.17}
	\setlength\tabcolsep{1.5mm}
	\centering
	%\begin{center}
	\makeatletter\def\@captype{table}\makeatother
	\caption{Candidate operations in our mobile search spaces.}
	\vspace{-2mm}
	\label{tab:op_mbv2}
	\footnotesize
	%\normalsize
	\begin{tabular}{c|c|c|c|c}
		\Xhline{2\arrayrulewidth}
		search space & block type & expansion ratio & kernel size & SE \\
		\hline
		- & MB1\_K3 & 1 & 3 & no\\
		\hline
		\multirow{13}*{MB-SE} & ID &- &-&- \\
		~ & MB3\_K3 & 3 & 3 & no \\
		~ & MB3\_K5 & 3 & 5 & no \\
		~ & MB3\_K7 & 3 & 7 & no \\
		~ & MB6\_K3 & 6 & 3 & no \\
		~ & MB6\_K5 & 6 & 5 & no \\
		~ & MB6\_K7 & 6 & 7 & no \\
		~ & MB3\_K3\_SE & 3 & 3 & yes \\
		~ & MB3\_K5\_SE & 3 & 5 & yes \\
		~ & MB3\_K7\_SE & 3 & 7 & yes \\
		~ & MB6\_K3\_SE & 6 & 3 & yes \\
		~ & MB6\_K5\_SE & 6 & 5 & yes \\
		~ & MB6\_K7\_SE & 6 & 7 & yes \\
		\hline
		\multirow{4}*{MixConv} & MB3\_MIX & 3 & 3+5+7 & no\\
		~ & MB6\_MIX & 6 & 3+5+7 & no\\
		~ & MB3\_MIX\_SE & 3 & 3+5+7 & yes\\
		~ & MB6\_MIX\_SE & 6 & 3+5+7 & yes\\
		\hline
		\multirow{4}*{Shuffle} & Shuffle\_3 & - & 3 & yes\\
		~ & Shuffle\_5 & - & 5 & yes\\
		~ & Shuffle\_7 & - & 7 & yes\\
		~ & Shuffle\_x & - & 3 + 3 + 3 & yes\\
		\Xhline{2\arrayrulewidth}
	\end{tabular}
	%\end{center}
\end{minipage}
\hfill
\begin{minipage}[c]{0.5\textwidth}
	\renewcommand\arraystretch{1.17}
	\setlength\tabcolsep{2mm}
	\centering
	%\begin{center}
	\makeatletter\def\@captype{table}\makeatother
	\caption{Candidate operations in our Res-50-SE search space.}
	\vspace{-2mm}
	\label{tab:op_res50}
	\footnotesize
	%\normalsize
	\begin{tabular}{l|c|c|c|c}
		\Xhline{2\arrayrulewidth}
        block type & basic block & ratio & kernel size & SE \\
        \hline
        ID &- &- &- & -\\
		\hline
		ResNet\_K3\_0.5$\times$ & ResNet & 0.5 & 3 & yes \\
		ResNet\_K3\_1$\times$ & ResNet & 1.0 & 3 & yes \\
		ResNet\_K3\_1.5$\times$ & ResNet & 1.5 & 3 & yes \\
		ResNet\_K5\_0.5$\times$ & ResNet & 0.5 & 5 & yes \\
    	ResNet\_K5\_1$\times$ & ResNet & 1.0 & 5 & yes \\
		ResNet\_K5\_1.5$\times$ & ResNet & 1.5 & 5 & yes \\
		ResNet\_K7\_0.5$\times$ & ResNet & 0.5 & 7 & yes \\
		ResNet\_K7\_1$\times$ & ResNet & 1.0 & 7 & yes \\
		ResNet\_K7\_1.5$\times$ & ResNet & 1.5 & 7 & yes \\
		\hline
		ResNet\_K3\_1$\times$ & ResNeXt & 1.0 & 3 & yes \\
		ResNeXt\_K3\_1.5$\times$ & ResNeXt & 1.5 & 3 & yes \\
		ResNeXt\_K5\_0.5$\times$ & ResNeXt & 0.5 & 5 & yes \\
		ResNeXt\_K5\_1$\times$ & ResNeXt & 1.0 & 5 & yes \\
		ResNeXt\_K5\_1.5$\times$ & ResNeXt & 1.5 & 5 & yes \\
		ResNeXt\_K7\_0.5$\times$ & ResNeXt & 0.5 & 7 & yes \\
		ResNeXt\_K7\_1$\times$ & ResNeXt & 1.0 & 7 & yes \\
		ResNeXt\_K7\_1.5$\times$ & ResNeXt & 1.5 & 7 & yes \\
		\Xhline{2\arrayrulewidth}
	\end{tabular}
	%\end{center}
\end{minipage}
\vspace{6mm}
\end{minipage}

\section{Implementing Details of PU Learning}

In the training of supernet, we train the path filter using VPU~\cite{chen2020vpu} every $t=5$ epoch. At the first time we train it, the weights of the path filter are randomly initialized, then the following training fine-tunes the weights obtained in the previous training.

\subsection{Complete learning objective in VPU}
The core idea of VPU is the proposed variational loss as in Eq.(6). Besides, to further alleviate the over-fitting problem, VPU incorporates a MixUp~\cite{zhang2018mixup} based consistency regularization term to the variational loss (Eq.(6)) as 
\begin{equation}
    \L_{\text{reg}}(\Phi) = \mathbb{E}_{\tilde{\Phi},\tilde{\a}} [(\mbox{log}\tilde{\Phi} - \mbox{log}\tilde{\Phi}(\tilde{\a}))^2] ,
\end{equation}
with
\begin{align}
    \begin{split}
        \gamma &\stackrel{\mathrm{iid}}{\sim} \mathrm{Beta}(\sigma, \sigma) ,\\
        \tilde{\a} &= \gamma \cdot \a' + (1 - \gamma) \cdot \a'', \\
        \tilde{\Phi} &= \gamma \cdot 1 + (1 - \gamma) \cdot \Phi(\a'') .
    \end{split}
\end{align}

Here $\tilde{\a}$ is an architecture generated by mixing randomly selected $\a'\in\P$ and $\a''\in\U$, and $\tilde{\Phi}$ represents the \textit{guessed} probability $\mathbb{P}(y=+1|\a=\tilde{\a})$ constructed by the linear interpolation of the true label and that predicted by $\Phi$, $\sigma$ is a hyper-parameter to control the MixUp percentage. Unlike the original MixUp on images, our architecture vector $\a$ is a tuple of discrete integers without semantic features, therefore, we conduct MixUp after the embedded features $\bm{A}$, \ie, 
\begin{equation}
    \bm{A}_{\tilde{\a}} = \gamma \cdot \bm{A}_{\a'} + (1 - \gamma) \cdot \bm{A}_{\a''} .
\end{equation}

\textbf{Complete form of loss function in VPU.} The complete loss function to train our path filter is as below:
\begin{equation}
    \L(\Phi) = \L_{\text{var}}(\Phi) + \lambda \L_{\text{reg}}(\Phi) .
\end{equation}
In our experiments, we set $\sigma = 0.3$ and $\lambda = 0.2$ following the original configurations in VPU.

\section{Searching Results of Baseline NAS Methods on Res-50-SE Search Space}

In this paper, we propose a new search space named \textit{Res-50-SE} for searching ResNet-like models. Here we conduct experiments to compare our method with baselines SPOS~\cite{guo2020single} and GreedyNAS~\cite{you2020greedynas}. As the results summarized in Table~\ref{tab:baselines_res}, we can see that our \textit{GreedyNASv2-L} obtains the highest accuracy with the minimal cost. Besides, for GreedyNAS, since each architecture has $\sim4$G FLOPs, the computation cost of multi-path sampling could be noticeably higher than the mobile search spaces, and the training cost is larger than GreedyNASv2 as a result.

\begin{table}[h]
	\renewcommand\arraystretch{1.2}
	\setlength\tabcolsep{3mm}
	\centering
	%\begin{center}
	\caption{Evaluation results of \textit{Res-50-SE} search space on ImageNet. The results of SPOS and GreedyNAS are obtained by our implementations.}
	\vspace{-2mm}
	\label{tab:baselines_res}
	\footnotesize
	%\normalsize
	\begin{tabular}{r|ccccc|ccc}
		\Xhline{2\arrayrulewidth}
		\multirow{2}*{Methods} & Top-1 & Top-5 & FLOPs & Params & Training & Training cost & Search\\
		%\cline{6-8}
		~ & (\%) & (\%) & (M) & (M) & epochs & (GPU days) & number\\
		\hline
		SPOS~\cite{guo2020single} & 80.6 & 95.1 & 4153 & 27.8 & 120 & 15.4 & 1000\\
		GreedyNAS~\cite{you2020greedynas} & 80.8 & 95.2 & 4125 & 28.1 & 49 & 11.3 & 1000 \\
		\textbf{GreedyNASv2-L} & \textbf{81.1} & 95.4 & 4098 & 26.9 & 57 & 9 & 500\\
		\Xhline{2\arrayrulewidth}
	\end{tabular}
	\vspace{-2mm}
	%\end{center}
\end{table}

\section{Transfer learning on object detection task}
We transfer our searched models to verify the generalization performance
on object detection task. Concretely, we train both two-stage Faster R-CNN
with Feature Pyramid Networks (FPN)~\cite{lin2017feature} and one-stage RetinaNet~\cite{lin2017focal} networks on COCO dataset~\cite{lin2014microsoft}, and report the validation mAP in Table~\ref{tab:detection}. Note that for fair comparisons, we train the networks using the default configurations in mmdetection~\cite{mmdetection}, with only modifications on backbone models. The results show that our obtained models significantly outperform the baseline models.

\begin{table}[h]
	\renewcommand\arraystretch{1.3}
	\setlength\tabcolsep{5mm}
	\centering
	%\begin{center}
	\caption{Evaluation results on COCO dataset.}
	\vspace{-2mm}
	\label{tab:detection}
	\footnotesize
	%\normalsize
	\begin{tabular}{c|c|c|c}
		\Xhline{2\arrayrulewidth}
		\multirow{2}*{Backbone} & ImageNet & FPN & RetinaNet\\
		%\cline{2-4}
        ~ & Top-1 (\%) & mAP (\%) & mAP (\%)\\
        \hline
        ResNet-50 & 76.1 & 37.4 & 36.5\\
        GreedyNASv2-L & 81.1 & 41.7 (\green{+4.3}) & 40.9 (\green{+4.4}) \\
        \hline
        MobileNetV2 & 72.0 & 32.1 & 30.5\\
        GreedyNASv2-S & 77.5 & 35.4 (\green{+3.3}) & 34.9 (\green{+4.4})\\
		\Xhline{2\arrayrulewidth}
	\end{tabular}
	\vspace{-2mm}
	%\end{center}
\end{table}

\section{More Ablation Studies}

\subsection{Effects of path-level shrinkage and operation-level shrinkage}
To validate the effectiveness of the proposed path-level shrinkage and operation-level shrinkage methods, we conduct experiments to ablate these two components in GreedyNASv2, as shown in Table~\ref{tab:effect_shrinkage}.

\begin{table}[h]
	\renewcommand\arraystretch{1.3}
	\setlength\tabcolsep{1.3mm}
	\centering
	%\begin{center}
	\caption{Effects of path-level and operation-level shrinkages on \textit{MB-SE+MixConv+Shuffle (large)} search space.}
	\vspace{-2mm}
	\label{tab:effect_shrinkage}
	\footnotesize
	%\normalsize
	\begin{tabular}{c|cc|cc}
		\Xhline{2\arrayrulewidth}
		Method & Path-level shrinkage & Operation-level shrinkage & ACC in retraining (\%) & ACC on supernet (\%)\\
		\hline
		SPOS~\cite{guo2020single} & - & - & 75.5 & 33.4\\
		GreedyNAS~\cite{you2020greedynas} & \faCheck & - & 76.5 & 35.1\\
		GreedyNASv2 & \faCheck & \faCheck & \textbf{77.5} & \textbf{43.8}\\
		\hline
		GreedyNASv2 & \faTimes & \faCheck & 76.8 & 42.1\\
		GreedyNASv2 & \faCheck & \faTimes & 77.2 & 39.6 \\
		\Xhline{2\arrayrulewidth}
	\end{tabular}
	\vspace{-2mm}
	%\end{center}
\end{table}

\subsection{Effect of path filter in search}
\begin{figure}[h]
	\centering
	%	\hspace{-1mm}	
	\subfigure[] 
	{\includegraphics[width=0.4\columnwidth]{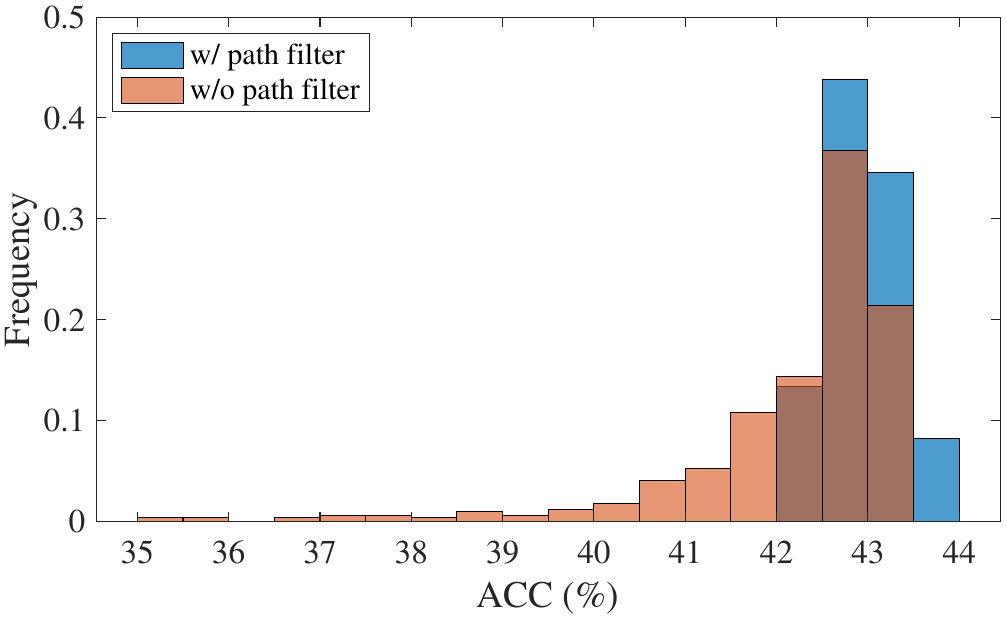}} 
	\hspace{10mm}
	\subfigure[]
	{\includegraphics[width=0.45\columnwidth]{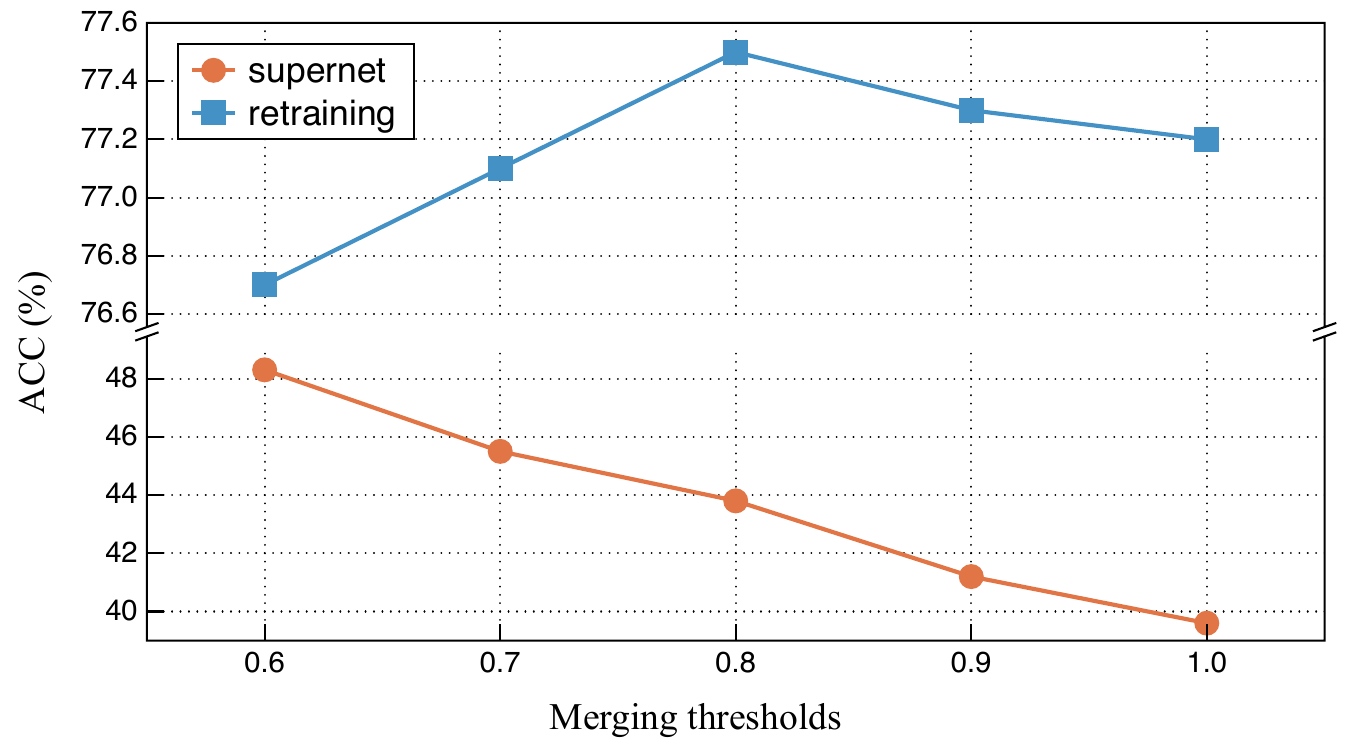}}
	\vspace{-2mm}
	\caption{(a): Histogram of accuracies of searched paths on supernet with or without using path filter. (b): Supernet and retraining accuracies of different operation merging thresholds. Specifically, the threshold of $1.0$ denotes no merging.}
	\label{fig:ab_acc}
\end{figure}

In GreedyNASv2, we use the learned path filter to filter the predicted weak paths during the search. Now we conduct experiments on \textit{MB-SE+MixConv+Shuffle} search space to show the effectiveness of path filter in search. As the histogram of searched accuracies shown in Figure~\ref{fig:ab_acc} (a), searching with a path filter can reject a larger number of weak paths and obtain higher accuracies, showing that the path filter can boost the searching efficiency.

\subsection{Effects of different operation merging thresholds}
In our operation-level shrinkage, the operation pair with a similarity large than a certain threshold would be treated as similar pair; then, we will merge them into one operation and obtain a smaller search space as a result. Here we conduct experiments to show the performance of different
merging thresholds. We train the \textit{MB-SE+MixConv+Shuffle} supernet using GreedyNASv2 with merging thresholds 0.6, 0.7, 0.8, 0.9, and 1.0, respectively, and report the supernet and retraining accuracies of the searched models in Figure~\ref{fig:ab_acc} (b). We can see that the smaller threshold would have more operations being merged, and thus the accuracy on supernet would be higher. However, too aggressive mergings (thresholds 0.6 and 0.7) would hurt the diversity of the search space; therefore, the performance of searched models would worsen.

\subsection{Validate the correctness of operation similarity} 
To validate the correlation between our learned operation similarities and their corresponding evaluation performance, we conduct experiments to measure the rank correlation of evaluation performance in each operation pair. Concretely, we split the operation pairs on \textit{MB-SE} search space into \textit{similar}, \textit{dissimilar}, and \textit{random} set, the \textit{similar} (\textit{dissimilar}) set contains $10$ pairs with highest (lowest) learned similarities, while the \textit{random} set are built with randomly generated pairs. We first measure the rank correlation of each pair independently, then report their mean correlation as the correlation of the set. Specifically, for the measurement of the rank correlation of each pair, we randomly generate $100$ paths containing the first operation, then evaluate their performance of validation set on a trained supernet, resulting in performance vector $\bm{x}$. For another operation, we use it to replace the first operation in generated paths and obtain performance vector $\bm{y}$. If these two operations in a pair have similar performance, vectors $\bm{x}$ and $\bm{y}$ will obtain similar values for each element. We then use Spearman's~\cite{dodge2008concise} and Kenall's Tau~\cite{kendall1938new} rank correlation to measure this similarity in performance. Note that we use the supernet learned by uniform sampling for fair evaluation without greedy biases.

As shown in Table~\ref{tab:correlation}, the learned similar pairs obtain very high similarities (rank correlations), indicating that our learned similarity can well reflect the similarity in performance; therefore, we can confidently leverage the learned similarities to merge operations.
\begin{table}[h]
	\renewcommand\arraystretch{1.17}
	\setlength\tabcolsep{3mm}
	\centering
	%\begin{center}
	\caption{Rank correlations of the evaluation results of similar, dissimilar, and random pairs identified by the path filter on supernet.}
	\vspace{-2mm}
	\label{tab:correlation}
	\footnotesize
	%\normalsize
	\begin{tabular}{c|c|c|c}
		\Xhline{2\arrayrulewidth}
		\multirow{2}*{Pairs} & \multirow{2}*{Mean similarity} & \multicolumn{2}{c}{Rank correlation (\%)}\\
		\cline{3-4}
		~ & ~ & Spearman's & Kendall's Tau\\
		\hline
		similar & 0.9118 & \textbf{99.26} & \textbf{93.68}\\
		dissimilar & -0.2183 & 73.38 & 69.58\\
		random & 0.3412 & 83.62 & 78.31\\
		\Xhline{2\arrayrulewidth}
	\end{tabular}
	\vspace{-4mm}
	%\end{center}
\end{table}

\newpage
\section{Visualization of our searched architectures}
\begin{figure}[h]
	\centering
	%	\hspace{-1mm}	
	\subfigure[GreedyNASv2-S] 
	{\includegraphics[width=0.18\columnwidth]{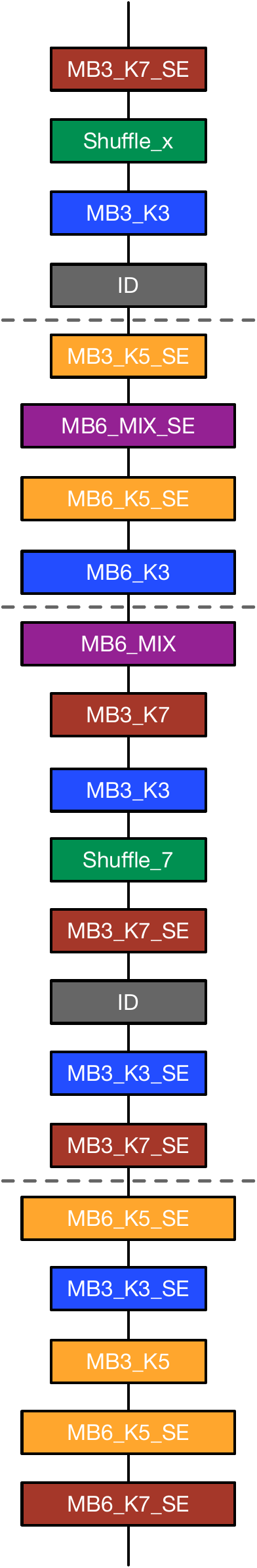}} 
	\hspace{35mm}
	\subfigure[GreedyNASv2-L]
	{\includegraphics[width=0.18\columnwidth]{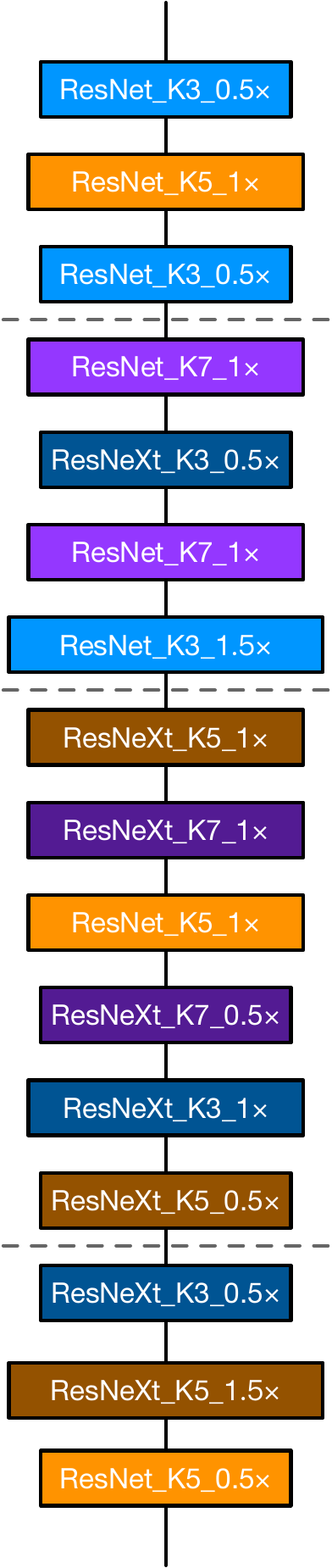}}
	\vspace{-2mm}
	\caption{Visualization of our searched architectures.}
	\label{fig:vis_model}
\end{figure}

Our searched \textit{GreedyNASv2-S} and \textit{GreedyNASv2-L} are visualized in Figure~\ref{fig:vis_model}. 

\end{document}